\documentclass[journal]{IEEEtran}

\usepackage{enumitem}

\usepackage[colorlinks,urlcolor=blue,linkcolor=blue,citecolor=blue]{hyperref}

\usepackage{color,array}

\usepackage{graphicx}

\ifCLASSINFOpdf
\else
\fi

\makeatletter
\newcommand*\bigcdot{\mathpalette\bigcdot@{.5}}
\newcommand*\bigcdot@[2]{\mathbin{\vcenter{\hbox{\scalebox{#2}{$\m@th#1\bullet$}}}}}
\makeatother

\usepackage{bm}
\usepackage{mathrsfs}
\usepackage{array}
\usepackage{amsfonts,amssymb}
\usepackage{siunitx}
\usepackage{amsmath}
\usepackage[ruled,linesnumbered]{algorithm2e}

\ifCLASSOPTIONcompsoc
\usepackage[caption=false,font=normalsize,labelfont=sf,textfont=sf]{subfig}
\else
\usepackage[caption=false,font=footnotesize]{subfig}
\fi

\usepackage{tabularx}
\usepackage[table,xcdraw]{xcolor}
\usepackage{multirow}
\usepackage{stfloats}

\usepackage{amsmath}

\usepackage{url}

\usepackage[colorlinks,linkcolor=blue]{hyperref}

\begin{document}

\title{ABCP: Automatic Block-wise and Channel-wise Network Pruning via Joint Search}

\author{Jiaqi Li, Haoran Li, \IEEEmembership{Member, IEEE}, Yaran Chen, \IEEEmembership{Member, IEEE},  Zixiang Ding, \IEEEmembership{Graduate Student Member, IEEE}, Nannan Li, Mingjun Ma, Zicheng Duan, and Dongbing Zhao, \IEEEmembership{Fellow, IEEE}
\thanks{This work is supported partly by the National Natural Science Foundation of China (NSFC) under Grants No. 62006226, the National Key Research and Development Program of China under Grant 2018AAA0101005, and the Strategic Priority Research Program of Chinese Academy of Sciences under Grant No. XDA27030400. (\textit{Corresponding author: Haoran Li and Yaran Chen}).}
\thanks{Jiaqi Li is with the State Key Laboratory of Management and Control for Complex Systems, Institute of Automation, Chinese Academy of Sciences, Beijing, 100190,
China, and also with the School of Mechanical Engineering, Beijing Institute of Technology, Beijing, 100081, China (email: xuer0324@gmail.com).}
\thanks{Haoran Li, Yaran Chen, Zixiang Ding, Nannan Li, Mingjun Ma and Dongbin Zhao are with the State Key Laboratory of Management and Control for Complex Systems, Institute of Automation, Chinese Academy of Sciences, Beijing, 100190, China, and also with the University of Chinese Academy of Sciences, Beijing, China (email: lihaoran2015@ia.ac.cn, chenyaran2013@ia.ac.cn, dingzixiang2018@ia.ac.cn, linannan2017@ia.ac.cn, mamingjun2020@ia.ac.cn, dongbin.zhao@ia.ac.cn).}
\thanks{Zicheng Duan is with the College of Engineering and Computer Science, Australian National University, ACT, 2601, Australia (email: zicheng.duan@anu.edu.au).}}
%\thanks{This paragraph will include the Associate Editor who handled your paper.}}

\markboth{Journal of IEEE Transactions on Cybernetics, Vol. 00, No. 0, Month 2021}
{Jiaqi Li \MakeLowercase{\textit{et al.}}: IEEE Journals of IEEE Transactions on Cybernetics}

\maketitle

\IEEEpeerreviewmaketitle

\begin{abstract}
Currently, an increasing number of model pruning methods are proposed to resolve the contradictions between the computer powers required by the deep learning models and the resource-constrained devices. However, most of the traditional rule-based network pruning methods can not reach a sufficient compression ratio with low accuracy loss and are time-consuming as well as laborious. In this paper, we propose Automatic Block-wise and Channel-wise
Network Pruning (ABCP\footnote[1]{Our code will be released at \href{https://github.com/DRL-CASIA/ABCP}{https://github.com/DRL-CASIA/ABCP}.}) to jointly search the block-wise and channel-wise pruning action with deep reinforcement learning. A joint sample algorithm
is proposed to simultaneously generate the pruning choice of each residual block and the channel pruning ratio of each convolutional layer from the discrete and continuous search space respectively. The best pruning action taking both the accuracy and the complexity of the model into account is obtained finally. Compared with the traditional rule-based pruning method, this pipeline saves human labor and achieves a higher compression ratio with lower accuracy loss. Tested on the mobile robot detection dataset, the pruned YOLOv3 model saves 99.5\% FLOPs, reduces 99.5\% parameters, and achieves $\bm{37.3\times}$ speed up with only 2.8\% mAP loss. The results of the transfer task on the sim2real detection dataset also show that our pruned model has much better robustness performance.
\end{abstract}


\section{Introduction}

\IEEEPARstart{I}{n} recent years, the deep learning methods are widely applied to machine learning tasks such as speech recognition \cite{speech}, image processing \cite{bifnet}, and structured output \cite{structure}. However, the large computational costs of the deep learning models are unaffordable for many resource-constrained devices and make the inference very slow.

To tackle these problems, a large number of approaches have been proposed \cite{heuristic} \cite{TCYBpruning}. Network pruning is a typical rule-based model compression method to reduce the redundant weights or structures in the network. There are two main types of network pruning: the non-structured pruning \cite{brain} and the structured pruning \cite{filters}. Since the structured pruning methods utilize structures as the pruning units, such as channels \cite{channel}, blocks \cite{block}, groups \cite{block} as well as both channels and blocks \cite{Compression}, so that the structured pruning is much more hardware-friendly than the non-structured pruning. Hence, most of the researchers tend to pay attention to structured model pruning currently, which is also the focus of this paper. 

Nevertheless, the existing structured pruning methods still have some problems. Firstly, fine-tuning the hyperparameters like the pruning threshold increases the workload and it is hard to prove which threshold is optimal \cite{channel} \cite{block}. Secondly, the traditional rule-based pruning method always cannot reach a sufficient compression ratio with low accuracy loss. Finally, the iterative pruning has been recommended \cite{Compression}\cite{uav}, making the ``sparse training – pruning – fine-tuning" pipeline be processed several times, which is time-consuming.

At present, the Neural Architecture Search (NAS) methods which can automatically search the network structure have gained ground \cite{nas} \cite{enas}. In this way, the networks with promising performance can be obtained efficiently with little human labor. To this end, we attempt to combine NAS and network pruning to achieve an automated process for the model pruning tasks. Some recent works have introduced the NAS methods into the model pruning procedure. Since AMC \cite{amc} utilized the deep reinforcement learning (DRL) to search each pruning ratio of each convolutional layer, several researchers realize the automated network pruning by DRL \cite{autocompress} or the evolutionary computation \cite{metapruning}. 

However, most of these works \cite{amc} only prune the channels and even can not prune the channels of the layers which belong to the residual blocks, making a very limited compression. The depth reduction of the network (e.g. residual block pruning) is also required to achieve high model pruning rates. Meanwhile, the discrete search space in some works \cite{abcpruner} also results in the confined compression ratio. In addition, the layers in the model are so sensitive to be pruned by different ratios that it is more suitable to utilize continuous search space.

Therefore, we propose Automatic Block-wise and Channel-wise Network Pruning (ABCP) to jointly search the channel-wise and block-wise pruning action by DRL. The action of DRL is a list consisting of the pruning choice of each residual block and the channel pruning ratio of each convolutional layer. The reward for DRL takes both the accuracy and complexity of the pruned model into account. Specifically, we propose a joint sample algorithm to generate the block-wise and channel-wise pruning action. We combine the discrete space and continuous space to sample the block pruning choice and the layer pruning ratio respectively. In addition, we also offer another choice to use the discrete space for sampling the layer pruning ratio. The joint sample algorithm is trained by the policy gradient method \cite{pg}.

Extensive experiments suggest that ABCP has a very promising performance. We collect three datasets, two of them are proposed by ourselves for the fast and lightweights detection algorithm for the mobile robots, and the third is captured from the videos collected by University of California San Diego (UCSD). Based on these three datasets, we evaluate ABCP on YOLOv3 \cite{yolov3}. Results show that our method achieves better accuracy than the traditional rule-based pruning method with fewer floating point operations (FLOPs). Furthermore, the results also demonstrate that the pruned model via ABCP has much better robustness performance.

To summarize, our contributions are listed as the following:
\begin{enumerate}[leftmargin=0.15in] 
\item[$\bullet$]We propose ABCP, a pipeline to jointly search the block-wise and channel-wise network pruning action by DRL.
\item[$\bullet$]We propose a joint sample algorithm to jointly sample the pruning choice of each residual block and the channel pruning ratio of each convolutional layer from the discrete and continuous space respectively.
\item[$\bullet$]We test ABCP on YOLOv3 \cite{yolov3} based on three datasets. The results show that ABCP outperforms the traditional rule-based pruning methods. On the mobile robot detection dataset, the pruned model saves 99.5\% FLOPs, reduces 99.5\% parameters, and achieves 37.3$\times$ speed up with only 2.8\% mAP loss. On the sim2real detection dataset, the pruned model has better robustness performance, achieving 9.6\% better mAP in the transfer task.
\end{enumerate}

\section{Related Works}

\subsection{Network Pruning}

Network pruning can be divided into non-structured pruning and structured pruning. Because of the sparse weight matrix, the acceleration of non-structured pruning in the hardware implementation is very limited. While the model compressed by the structured pruning is easier to be deployed on the hardwares.

The structured pruning method takes the structures as the basic pruning unit, such as channels, residual blocks, and residual groups. Li et al. \cite{filters} firstly proposed the method to prune the unimportant filters of the convolutional layers, evaluating the importance of each filter via the absolute sum of the weights. Liu et al. \cite{channel} presented a ``sparse learning – pruning – fine-tuning" pipeline to prune channels. He et al. \cite{asymptotic} developed a soft channel pruning pipeline to remedy the problem of information loss in \cite{channel}.  Then, a method for pruning residual blocks and residual groups was proposed by Huang and Wang \cite{block}. Zhang, Zhong and Li \cite{uav} extended the pruning method to the detection tasks, which iteratively pruned YOLOv3 to accelerate the model on unmanned aerial vehicles. Then, Li et al. \cite{Compression} combined the block-wise pruning and channel-wise pruning of YOLOv3 by elaborately designed rules to reduce the cost of the model on the environment perception devices of vehicles. In this paper, we also aim to automatically prune blocks and channels of the YOLOv3 \cite{yolov3} model via joint search.

\subsection{Automatic Network Pruning}
Currently, Neural Architecture Search (NAS) methods have been proposed to automatically search network architectures to reduce the intensity of human labor \cite{bnas} \cite{modulenet}. Then, several papers have presented the techniques to combine network pruning and NAS. AMC \cite{amc} proposed an Automatic Machine Learning (AutoML) engine to generate the pruning ratio of each layer with continuous search space. MetaPruning \cite{metapruning} utilized the evolutionary computation to search the pruned networks, and then the structure and weights can be efficiently sampled from a meta network without fine-tuning. DMCP \cite{dmcp} introduced a differentiable search method for channel pruning, with modeling the pruning procedure as a Markov process. CACP \cite{cacp} also cast the channel pruning into a Markov decision procedure, and the pruned models with different compression ratios can be searched at the same time. ABCPruner \cite{abcpruner} optimized the pruned structure by the artificial bee colony algorithm, but the search space of the pruning was discrete. AACP \cite{aacp} presented an automatic channel pruning algorithm that can optimize the FLOPs, inference time, and model size simultaneously. Nevertheless, these methods only considered channel-wise pruning. AutoCompress \cite{autocompress} combined the channel pruning and the non-structured pruning and searched the best pruning action by DRL. In this paper, through DRL with both discrete and continuous search space, we aim to search for a block-wise and channel-wise pruning action that can reduce resource costs with almost no accuracy loss.

\section{Methodology}

\begin{figure*}[t]
\centering
\includegraphics[width=5.5in]{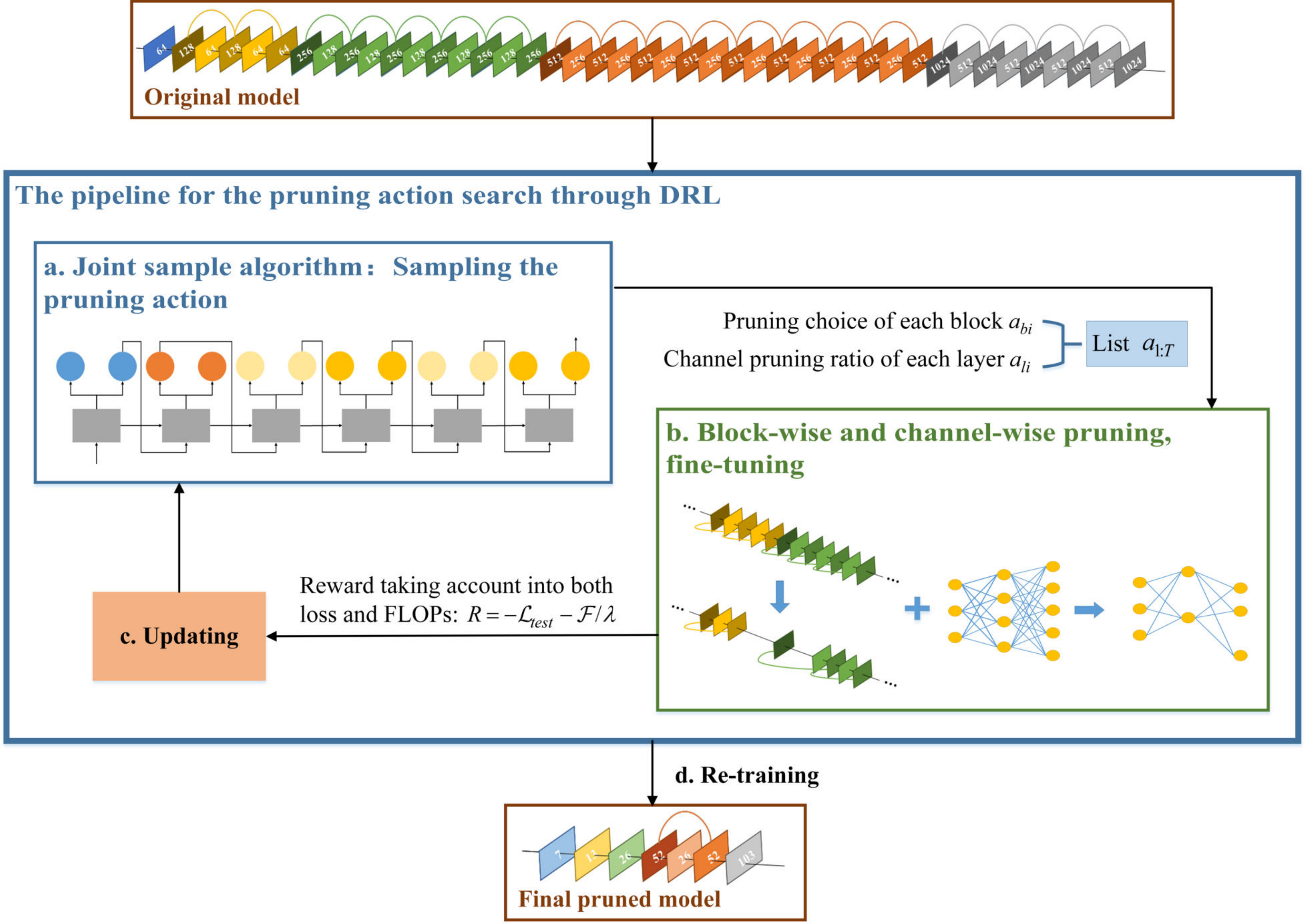}
\caption{The framework of Automatic Block-wise and Channel-wise Network Pruning (ABCP). The input of this pipeline is a large pre-trained residual network and the final pruned model has much fewer residual blocks and channels. In each episode, the first step is using the joint sample algorithm to sample the pruning action $a_{1:T}$ which includes the pruning choice of each residual block $a_{bi}$ and the channel pruning ratio of each convolutional layer $a_{li}$, and then the model is pruned by setting the corresponding weights to zero according to the sampled pruning action. After pruning, several particular layers in the pruned model is fine-tuned on the training dataset with maintaining the values of the weights that have been set to zero in the previous step. In the final step, the loss on the testing dataset $\mathcal{L}_{test}$ is calculated and the FLOPs of the pruned model $\mathcal{F}$ is estimated to generate the reward $R$, then the weights of the joint sample algorithm are updated by the policy gradient method.}
\label{fig1}
\end{figure*}

In this section, we present our pruning action search method ABCP in Fig. \ref{fig1}. The method aims to automatically search the block-wise and channel-wise redundancy of the overall model by DRL. Inspired by the methods for neural network search \cite{nas} \cite{enas}, for the network which has $T$ layers to prune, a list $a_{1:T}$ consisting of the pruning choice of each block $a_{bi}$ and the channel pruning ratio of each layer $a_{li}$ can be considered as the action for DRL. After pruning and fine-tuning, the testing loss $\mathcal{L}_{test}$ and the FLOPs of the pruned network $\mathcal{F}$ can be used as the reward for DRL. Specifically, a joint sample algorithm which utilizes a stack long short-term memory (LSTM) \cite{lstm} network has been proposed to the generate block-wise and channel-wise pruning action. The details of the framework are elaborated as follows:

\begin{enumerate}[leftmargin=0.18in]
\item[a)]Sampling the pruning action: The representation of the large pre-trained model is fed into the joint sample algorithm, and then the pruning action including the block pruning choice for each residual block and the channel pruning ratio for each convolutional layer is sampled. (see Sec.\uppercase\expandafter{\romannumeral3}.A)
\item[b)]Pruning and fine-tuning: Once the pruning action is generated, the corresponding weights in the original models are set to zero. With maintaining the values of the weights which have been set to zero, several particular layers in the model is fine-tuned on the training dataset. (see Sec.\uppercase\expandafter{\romannumeral3}.B)
\item[c)]Updating: After finishing the fine-tuning, the loss of the pruned model is calculated on the testing dataset and the FLOPs of the pruned model is estimated. Then the reward that takes both accuracy and FLOPs into account is calculated, and the parameters of the joint sample algorithm are updated by the policy gradient method. (see Sec.\uppercase\expandafter{\romannumeral3}.C)
\item[d)]Re-training: After several episodes, the pruning action with the best reward is selected, and the final pruned model is re-trained from scratch. (see Sec.\uppercase\expandafter{\romannumeral3}.D)
\end{enumerate}

\subsection{Sampling the Pruning Action}

\subsubsection{The joint sample algorithm}

\begin{figure*}[t]
    \centering
    \subfloat[The model of the joint sample algorithm.]{\includegraphics[width=7.2in]{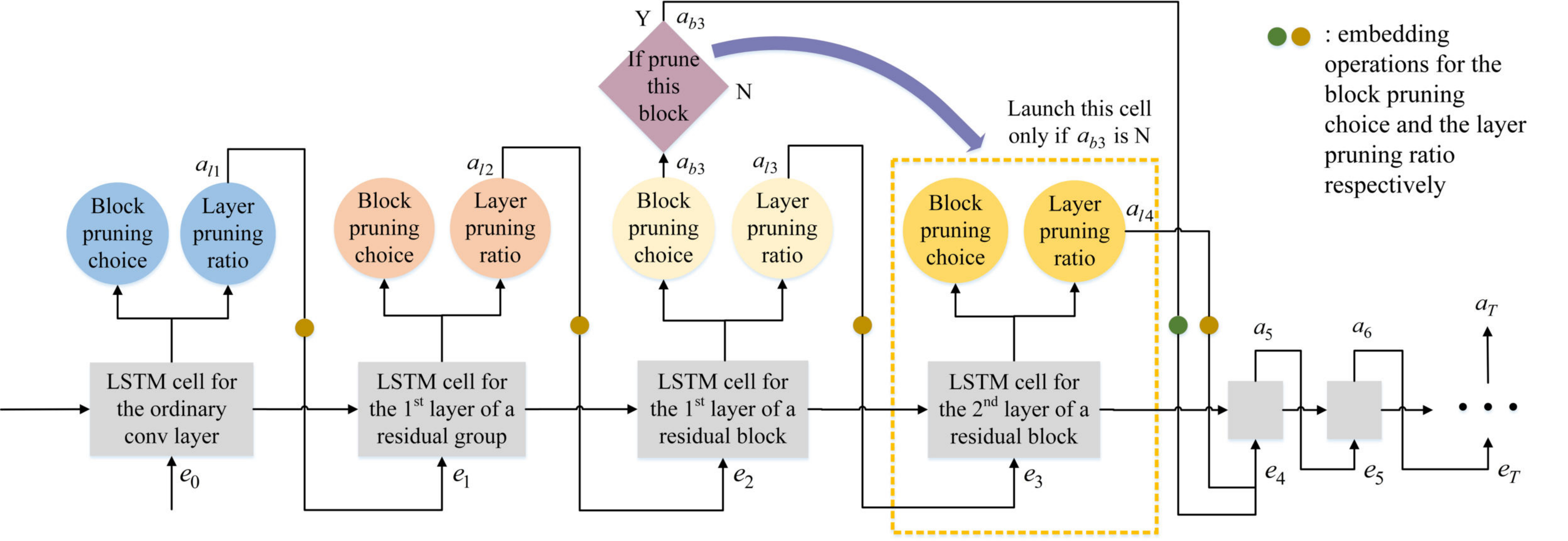}}
    \vspace{5mm}
    \hfil
    \subfloat[The structure of the residual network that will be pruned.]{\includegraphics[width=7.2in]{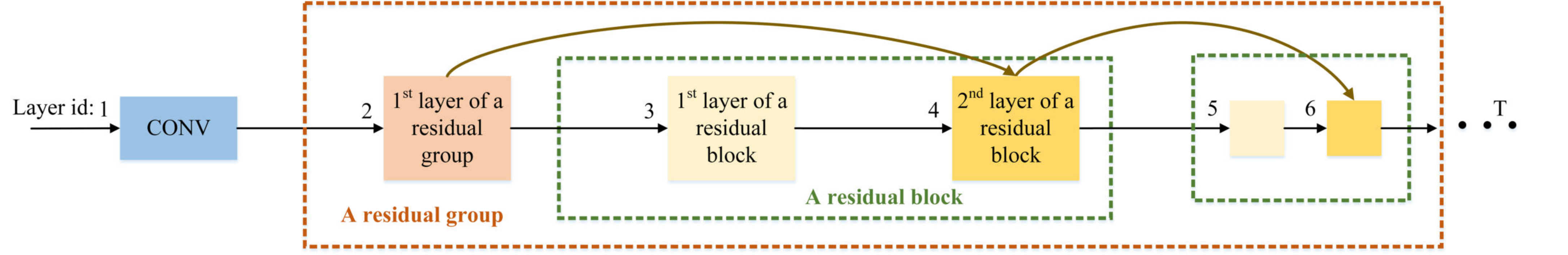}}
    \vspace{5mm}
    \caption{The process of the pruning action sampling. (a) The model of the joint sample algorithm; (b) The structure of the residual network that will be pruned. The LSTM network is used for the joint sample algorithm where each cell is connected by two branches: one branch outputs the block pruning choice and the other branch outputs the layer pruning ratio. We denote four types of the layers in this residual network, so there are also four types of LSTM cells: the LSTM cell for the ordinary convolutional layer, the LSTM cell for the 1st layer of a residual block, the LSTM cell for the 2nd layer of a residual block, and the LSTM cell for the 1st layer of a residual group. During the pruning action sampling, each LSTM cell samples the corresponding block pruning choice $a_{bi}$ or the layer pruning ratio $a_{li}$ for each block and each layer in the residual network respectively, which constitutes the list $a_{1:T}$ as the pruning action for the whole residual network.}
    \label{fig2}
    \end{figure*}

As demonstrated in Fig. \ref{fig2}, to sample the block-wise and channel-wise pruning action, we propose a joint sample algorithm using the LSTM model. The structure of the residual network that would be pruned  is also shown in Fig. \ref{fig2}, involving an ordinary convolutional layer and a residual group that consists of two residual blocks, each residual block consists of two convolutional layers. Each LSTM cell samples the block pruning choice $a_{bi}$ or the layer pruning ratio $a_{li}$ for each corresponding block or layer of the residual network respectively, which constitutes the list $a_{1:T}$ as the pruning action of the residual network.

Fig.\ref{fig2} also illustrates that there are two branches connected with each LSTM cell to sample the block-wise and channel-wise pruning action. Each branch consists of one or two fully connected (FC) layers and the softmax operation. One branch is to sample the pruning choice of each residual block $a_{bi}$, and the other is to sample the channel pruning ratio of each convolutional layer $a_{li}$. In addition, following \cite{enas}, the block pruning choice or the layer pruning ratio sampled in the previous cell are embedded in the next cell as the input $e_i$. In this way, a continuous distributed representation is created to capture the potential relationships between the current situation of the pruned network and the block pruning choices as well as the layer pruning ratios sampled by the previous LSTM cells. 

Especially, there are four types of LSTM cells according to the positions of the layers they control: the LSTM cell for the ordinary convolutional layer, the LSTM cell for the 1st layer of a residual block, the LSTM cell for the 2nd layer of a residual block, and the LSTM cell for the 1st layer of a residual group:

\begin{enumerate}[leftmargin=0.15in] 
    \item[$\bullet$]The LSTM cell for the ordinary convolutional layer: The LSTM cell for the ordinary convolutional layer takes the sampled layer pruning ratio as the output of this cell, then embeds and feeds it into the next cell along with the cell state and the recurrent information.
    \item[$\bullet$]The LSTM cell for the 1st layer of a residual block: The LSTM cell for the 1st layer of the residual block makes the block pruning choice whether pruning both the two layers involved in this block or not. When the choice is Yes, the block pruning choice is embedded and fed into the cell after the next cell. When the sampled block pruning choice is No, this cell outputs are the sampled results of the layer pruning ratio branch, which is embedded and fed into the next cell.
    \item[$\bullet$]The LSTM cell for the 2nd layer of a residual block: Whether to launch the LSTM cell for the 2nd layer of a residual block depends on the sampled block pruning choice of the previous cell.
    \item[$\bullet$]The LSTM cell for the 1st layer of a residual group: As for the LSTM cell for the 1st layer of a residual group, since the 1st layers of the residual groups usually include pooling operations, we treat these layers as the ordinary convolutional layers to only prune the channels for maintaining the accuracy performance.
    \end{enumerate}

\subsubsection{Sampling the block pruning choice}

For the block pruning choice search, the action space is discrete, including ``pruning" and ``no pruning", i.e. ${{a_{bi}} \in \{1, 0\}}$. The probabilities of ``pruning" and ``not pruning" are computed by the softmax operation. Then the joint sample algorithm samples the block pruning choice from the probability distribution. The sampling process of the $i$th LSTM cell is denoted as:
\begin{eqnarray}
\bm{B_i} &=& \mathcal{B}_i(e_i;\bm{{\theta}}_{bi})\label{eq1} \\
P({a_{bi}}={\mathbb{B}_k}) &=& \frac{{\exp (B_i^k)}}{{\sum {_{{k'} = 1}^K\exp (B_i^{{k'}})} }}\label{eq2} \\
{a_{bi}} &\sim& {\pi({a_{bi}} \in \mathbb{B}|s; \bm{{\theta}}_{bi})}\label{eq3}
\end{eqnarray}
where $\mathbb{B}$ is the action space for the block pruning choice search, $K$ is the cardinality of $\mathbb{B}$, i.e. $K=|\mathbb{B}|$, $K=2$ here; $\mathbb{B}_k$ is the $k$th element in $\mathbb{B}$; $\mathcal{B}_i$ in \eqref{eq1} is the FC layer of the block pruning choice branch connected with the $i$th LSTM cell, $e_i$ is the embedded result of the $i$-1th block pruning choice, and $\bm{{\theta}}_{bi}$ denotes the weights of $\mathcal{B}_i$; $\bm{B_i}$ is the output of $\mathcal{B}_i$ with $K$ dimensions, and $B_i^k$ is the $k$th element of $\bm{B_i}$; $a_{bi}$ is the block pruning choice sampled by the $i$th LSTM cell, $s$ is the current state. \eqref{eq2} illustrates the softmax calculating process, which generates the probability distribution of $a_{bi}$. Then, as denoted in \eqref{eq3}, $a_{bi}$ is sampled from the distribution $\pi({a_{bi}} \in \mathbb{B}|s; \bm{{\theta}}_{bi})$.

After sampling, a embedding map is learned and then each block pruning choice in the discrete action space is mapped to a tensor in the continuous vector space through the embedding layer.

\subsubsection{Sampling the layer pruning ratio}

For the layer pruning ratio search, the action space can be discrete or continuous. The discrete action space is a coarse-grained space, i.e. ${{a_{li}} \in \{0, 0.225, 0.45, 0.675, 0.9\}}$. The probability of each pruning ratio is calculated by the softmax operation. However, it is sensitive for the layers to be pruned by different pruning rates, so that the fine-grained search space may be more suitable for the layer pruning ratio search.  Therefore, we also introduce the continuous action space to guide more fine-grained channel pruning, i.e. ${{a_{li}} \in [0, 0.9]}$.

For the discrete action space, the sampling process is similar to that of the block pruning choice:
\begin{eqnarray}
\bm{D_i} &=& \mathcal{D}_i(e_i;\bm{{\theta}}_{li})\label{eq4} \\
P({a_{li}}=\mathbb{D}_m)&=&\frac{{\exp (D_i^m)}}{{\sum {_{{m'} = 1}^M\exp (D_i^{{m'}})} }} \label{eq5} \\
{a_{li}} &\sim& \pi({a_{li}} \in \mathbb{D}|s;\bm{{\theta}}_{li})\label{eq6}
\end{eqnarray}
where $\mathbb{D}$ is the discrete action space for the layer pruning ratio search, $M$ is the cardinality of $\mathbb{D}$, i.e. $M=|\mathbb{D}|$, $M=5$ here according to the action space mentioned above; $\mathbb{D}_m$ is the $m$th element of $\mathbb{D}$; $\mathcal{D}_i$ in \eqref{eq4} is the FC function of the layer pruning ratio branch connected with the $i$th LSTM cell, $e_i$ is the embedded result, and $\bm{{\theta}}_{li}$ denotes the weights of $\mathcal{D}_i$; $\bm{D_i}$ is the output of $\mathcal{D}_i$ with $M$ dimensions, and $D_i^m$ is the $m$th element of $\bm{D_i}$; $a_{li}$ is the layer pruning ratio sampled by the $i$th LSTM cell. \eqref{eq5} demonstrates the softmax calculating process, which can obtain the probability distribution of $a_{li}$. Finally, \eqref{eq6} shows that $a_{li}$ is sampled from the distribution $\pi({a_{li}} \in \mathbb{D}|s; \bm{{\theta}}_{li})$.

For the continuous action space, we use the Gaussian distribution to represent the distribution of the layer pruning ratio and sample the ratio with the distribution. The $i$th LSTM cell is connected with two FC layers to generate the mean and log variance of the Gaussian distribution respectively.
\begin{eqnarray}
{\hat{\mu_i}} &=& \bm{\mu}_i(e_i; \bm{{\theta}}^{\mu}_{li})\label{eq7}  \\
{\hat{\rho_i}} &=& \bm{\rho}_i(e_i;  \bm{{\theta}}^{\rho}_{li})\label{eq8}  \\
{\hat{\sigma_i}}^2 &=& \exp(\hat{\rho_i})\label{eq9} \\
\pi(a_{li}|s;\bm{{\theta}}^{\mu}_{li},\bm{{\theta}}^{\rho}_{li}) &\sim& \mathcal{N}({\hat{\mu_i}},{{\hat{\sigma_i}}^2})\label{eq10} \\
{a_{li}} &\sim& \pi({a_{li}} \in \mathbb{C}|s;\bm{{\theta}}^{\mu}_{li}, \bm{{\theta}}^{\rho}_{li})\label{eq16}
\end{eqnarray}
where $\bm{\mu}_i$ and $\bm{\rho}_i$ in \eqref{eq7} and \eqref{eq8} are the two FC layers in the $i$th LSTM cell that estimate the mean $\hat{\mu_i}$ and log variance $\hat{\rho_i}$ of the Gaussian distribution respectively, ${\hat{\sigma_i}}^2$ is the variance of the distribution, and $\bm{{\theta}}^{\mu}_{li}$ and $\bm{{\theta}}^{\rho}_{li}$ are the weights of $\bm{\mu}_i$ and $\bm{\rho}_i$. Experiments show that approximating the log variance has a better practice, so we set the output of the FC layer $\hat{\rho_i}$ as the log variance. Then the variance ${\hat{\sigma_i}}^2$ is generated with $\hat{\rho_i}$, as shown in \eqref{eq9}. \eqref{eq10} illustrates that the distribution of the layer pruning ratio ${a_{li}}$ can be the Gaussian distribution $\mathcal{N}({\hat{\mu_i}},{{\hat{\sigma_i}}^2})$. \eqref{eq16} shows that ${a_{li}}$ is sampled in the action space $\mathbb{C}$ (i.e. $[0,0.9]$ here) from the distribution $\pi({a_{li}} \in \mathbb{C}|s;\bm{{\theta}}^{\mu}_{li}, \bm{{\theta}}^{\rho}_{li})$.

After sampling, the layer pruning ratio sampled by each cell are embedded. For the discrete action space, each pruning ratio is mapped to a tensor in the continuous vector space through an embedding layer and fed into the next cell. For the continuous search space, the pruning ratio is first rounded down, and then is mapped to a tensor.

The joint sample algorithm is detailed in Algorithm\ref{alg:algorithm1}.

\begin{algorithm}[!t]
	\caption{The joint sample algorithm}
	\label{alg:algorithm1}
	\KwIn{the LSTM model $\bm{\mathcal{C}}$ with $T$ cells denoted as $\mathcal{C}_{1},\mathcal{C}_{2},...,\mathcal{C}_{T}$; the branch that outputs the block pruning choice of the $i$th each cell: $\mathcal{C}_{bi}$; the branch that outputs the layer pruning ratio of the $i$th each cell: $\mathcal{C}_{li}$; the cell state, recurrent information, and embedded result inputted in the $i$th cell: $c_{i-1}$, $h_{i-1}$, $e_{i-1}$; the set of the ids of the first layer in each residual block in the original network: $\mathcal{S}_{frb}$.}

	\BlankLine

    Initialize the pruning action as an empty list ${a}_{1:T}$. ${a}_{1:T} = [\quad]$;

    Initialize ${c}_{0}$, ${h}_{0}$, ${{e}_{0}}$;

    %${c}_{1}$, ${h}_{1}$  $\gets$ $\mathcal{C}_{1}$(${{e}_{0}}$, ${c}_{0}$, ${h}_{0}$)

    %\eIf{${1}$ \textnormal{in} $\mathcal{S}_{frb}$}{
        %Generate the block pruning choice probability distribution by branch $\mathcal{C}_{b1}$: \eqref{eq2} $\gets$ $\mathcal{C}_{b1}$,

        %Sample ${a}_{b1}$ with \eqref{eq3};

        %\uIf{${a}_{b1}$=${0}$}{
             %Generate the layer pruning ratio probability distribution by branch $\mathcal{C}_{l1}$: \eqref{eq10} $\gets$ $\mathcal{C}_{l1}$,

            %Sample ${a}_{l1}$ with \eqref{eq16},
            %${a}_{1}$ = ${a}_{l1}$;}
        %\lElse{${a}_{1}$ = ${a}_{b1}$}

    %}{
     %\eqref{eq10} $\gets$ $\mathcal{C}_{l1}$,

     %Sample ${a}_{l1}$ with \eqref{eq16}, ${a}_{1}$ = ${a}_{l1}$;}

    %Add ${a}_{1}$ into ${a}_{1:n,t}$;

	\For{${i=1...T}$}{
        \lIf{${i}\neq 1$}{Embed ${a}_{i-1}$ as ${{e}_{i-1}}$}
    
		\uIf{${i}$ \textnormal{in} $\mathcal{S}_{frb}$}{

        ${c}_{i}$, ${h}_{i}$  $\gets$ $\mathcal{C}_i$(${{e}_{i-1}}$, ${c}_{i-1}$, ${h}_{i-1}$),

        Generate the block pruning choice probability distribution by branch $\mathcal{C}_{bi}$: \eqref{eq2} $\gets$ $\mathcal{C}_{bi}$,

        Sample ${a}_{bi}$ with \eqref{eq3};

            \uIf{${a}_{bi}$=${0}$}{
                Generate the layer pruning ratio probability distribution by branch $\mathcal{C}_{li}$: \eqref{eq10} $\gets$ $\mathcal{C}_{li}$,

                Sample ${a}_{li}$ with \eqref{eq16}, ${a}_{i}$ = ${a}_{li}$;}

            \lElse{${a}_{i}$ = ${a}_{bi}$}
        }
        \uElseIf{${i-1}$ \textnormal{in} $\mathcal{S}_{frb}$}{
            \uIf{${a}_{bi-1}$=${0}$}{
                ${c}_{i}$, ${h}_{i}$  $\gets$ $\mathcal{C}_i$(${{e}_{i-1}}$, ${c}_{i-1}$, ${h}_{i-1}$),

                \eqref{eq10} $\gets$ $\mathcal{C}_{li}$,

                Sample ${a}_{li}$ with \eqref{eq16}, 
                ${a}_{i}$ = ${a}_{li}$;}
            \lElse{${a}_{i}$ = ${a}_{bi-1}$, ${c}_{i}$ = ${c}_{i-1}$, ${h}_{i}$ = ${h}_{i-1}$
                }
        }
        \uElse{
            ${c}_{i}$, ${h}_{i}$  $\gets$ $\mathcal{C}_i$(${{e}_{i-1}}$, ${c}_{i-1}$, ${h}_{i-1}$),

            \eqref{eq10} $\gets$ $\mathcal{C}_{li}$,

            Sample ${a}_{li}$ with \eqref{eq16},
            ${a}_{i}$ = ${a}_{li}$;}

        Add ${a}_{i}$ into ${a}_{1:T}$;
	}
	
	\KwOut{the pruning action $a_{1:T}$.}
\end{algorithm}

\subsection{Pruning and Fine-tuning}

Once the pruning action is generated, the original pre-trained model should be pruned and fine-tuned. For the block pruning, we directly set the weights of the layers involved in the blocks to zero. Due to the existence of the shortcut operations, the block pruning does not influence the inference of the network. As for the channel pruning, we are supposed to select which channel to prune firstly.

As proposed in \cite{channel} and \cite{smallernorm}, the absolute value of the scale factor $\gamma$ in the batch normal (BN) layer can represent the importance of the channel. BN layer follows the convolutional layer in the network, which can be formulated as:
\begin{equation}
x_{out}^{p,q} = {\gamma ^{p,q}}\frac{{x_{in}^{p,q} - \phi _{\Omega}^{p,q}}}{{\sqrt {\delta _{\Omega}^{p,q} + \varepsilon } }} + {\beta ^{p,q}} \label{eq15}
\end{equation}
where the superscript ${p,q}$ means the ${q}$th channel of the ${p}$th convolutional layer; ${x_{in}^{p,q}}$ and ${x_{out}^{p,q}}$ are the input and output of the BN layer; ${\phi _{\Omega}^{p,q}}$ and ${\delta_{\Omega}^{p,q}}$ are the mean and standard deviation of $x_{in}^{p,q}$ over the ${\Omega}$th batch; ${\gamma ^{p,q}}$ and ${{\beta ^{p,q}}}$ are the scale and shift parameters, which are trainable.

Since ${x_{in}^{p,q}}$ is the output of the ${q}$th channel of the ${p}$th convolutional layer, ${\gamma ^{p,q}}$ determines the output of the corresponding channel. In addition, ${x_{in}^{p,q}}$ has been normalized to multiply with ${\gamma ^{p,q}}$. So the importance of the channel can be represented by the absolute value of ${\gamma ^{p,q}}$. Hence, we sort the absolute values of ${\gamma}$ in the original model and set the weights of the channels with smaller absolute ${\gamma}$ to zero according to the layer pruning ratios.

It is worth noting that the numbers of the channels in the convolutional layers connected by the shortcut operations in each residual block must be equal. To this end, all of the pruning ratios of these layers are forced to be equal to the maximum pruning ratio in the residual groups.

After pruning, we fine-tune the pruned model for one epoch to recover the accuracy with maintaining the values of the weights that have been set to zero. Especially, to speedup the pruning action search, as mentioned in Sec.\uppercase\expandafter{\romannumeral4}.B, we only fine-tune several particular layers.

\subsection{Updating}

The parameters of the joint sample algorithm are updated by the policy gradient method. Firstly, as shown in \eqref{eq11}, we define a reward $R$ that takes both the accuracy evaluated on the testing dataset and the cost of the pruned model into account for assessing the joint sample algorithm performance:
\begin{equation}\label{eq11}
R = - \mathcal{L}_{test} - \mathcal{F} / \lambda
\end{equation}
where $\mathcal{L}_{test}$ is the loss of the pruned and fine-tuned model, which is calculated on the testing dataset; $\mathcal{F}$ is the estimated total FLOPs of the pruned model; $\lambda$ is a trade-off hyper-parameter to balance the accuracy and the complexity. During the joint sample algorithm updating, we compute the sum of the FLOPs of every convolutional layer $\mathcal{F}_{layer}$ to approximate the total FLOPs $\mathcal{F}$ of the pruned model. \eqref{eq12} demonstrates how to estimate $\mathcal{F}_{layer}$:
\begin{equation}\label{eq12}
\mathcal{F}_{layer} = H \times W \times S \times S \times {C_{in}} \times {C_{out}}
\end{equation}
where $H$ and $W$ are the height and width of the feature map input in the convolutional layer, $S$ is the kernel size, $C_{in}$ and $C_{out}$ are the numbers of the input channels and the output channels of the convolutional layer respectively.

As shown in \eqref{eq17}, the goal of the policy gradient method is to maximize the expected reward to find the best pruning action, represented by $J(\theta)$:

\begin{equation}\label{eq17}
J(\theta) = {\mathbb{E}_{{\pi}(a_{1:T};\bm{\theta})}}[R(a_{1:T})]
\end{equation}
Where the reward $R(a_{1:T})$ is computed with the pruning action $a_{1:T}$ for the model which have $T$ layers to prune, and $a_{1:T}$ is sampled from the probability distribution ${{\pi}(a_{1:T};\bm{\theta})}$; $\bm{\theta}$ denotes the weights of the joint sample algorithm. 

We employ the Adam optimizer \cite{adam} to optimize the parameters of the joint sample algorithm, and the gradient is computed by REINFORCE \cite{pg} as \eqref{eq14}:
\begin{equation}\label{eq14}
{\nabla _\theta }J(\theta ) = \sum\limits_{t = 1}^T \mathbb{E}_{{\pi}({a_{1:T}};\bm{\theta})} [{\nabla _\theta }\log \pi({a_{t}}|s;\bm{\theta}){R(a_{1:T})}]
\end{equation}
where $s$ is the current state, which can be denoted as ${a_{(t-1):1}}$ in this task.

Through the Monte Carlo estimate, an empirical approximation of \eqref{eq14} is shown as \eqref{eq13}. And in order to reduce the high variance, a moving average baseline is employed in this estimate: 
\begin{equation}\label{eq13}
{\nabla _\theta }J(\theta ) \approx \frac{1}{N} \sum\limits_{n = 1}^N \sum\limits_{t = 1}^T {\nabla _\theta }\log \pi({a_{t}}|{a_{(t-1):1}};\bm{\theta})[{R(a_{1:T,n})-b}]
\end{equation}
where $N$ is the number of different pruning actions $a_{1:T}$ that the joint sample algorithm samples in one episode; ${R(a_{1:T,n})}$ is the reward calculated with the $n$th pruning action; $b$ is the moving average baseline. According to \cite{enas}, although the Monte Carlo estimate can approximate ${\nabla _\theta }J(\theta )$ unbiasedly, this method would bring a high variance when only the joint sample algorithm is trained. Furthermore, \cite{enas} has found that $N=1$ works well. So we let $N=1$ and let the reward calculated with one pruning action sampled from ${{\pi}(a_{1:T};\bm{\theta})}$ be the expected reward.

\subsection{Re-training}

After finishing the joint sample algorithm training and searching, several candidate pruning actions are obtained. We only take the pruning action with the highest reward to get a new pruned architecture, then the pruned model is re-trained from scratch. Maybe it is better for us to prune and re-train the models pruned by all the sampled pruning actions and to choose the one with the best performance, but it is time-consuming.

\section{Experiments}
In the area of computer vision, object detection plays an essential role and is applied in increasing industrial areas. Among the existing detectors, YOLOv3 \cite{yolov3} is the most promising and classic model with excellent real-time performance and considerable accuracy. Extensive diversity of methods have been presented to improve YOLOv3 (e.g. YOLOv4 \cite{yolov4}, etc.), while compared with these new modified networks, the operations in YOLOv3 are much more basic and simple, which are easier to be put into use. Currently, the YOLOv3 model is widely deployed on the embedded devices, hence the hope is to achieve better performance to meet the practical needs by compression.

To this end, YOLOv3 is adopted to illustrate the performance of our proposed ABCP framework in this section. We first introduce three datasets namely the UCSD dataset, the mobile robot detection dataset, as well as the sim2real detection dataset respectively, and then present the implementation settings. Secondly, the ablation experiments on the UCSD dataset are conducted to learn the significance of the block-wise and channel-wise joint search and the continuous action space of the search for channel pruning. Next, the effectiveness of our pruned model is evaluated on the UCSD dataset and the mobile robot detection dataset. Finally, the robustness of our method is demonstrated on the sim2real detection dataset.

\subsection{Dataset}
The series of the YOLO models are designed for relatively complex detection tasks, such as the object detection tasks on the VOC dataset (20 classes) \cite{voc} and the COCO dataset (80 classes) \cite{coco}. However, in many practical applications, the abundant detection classes are not needed since the objects in the detection tasks are relatively single, while the real-time requirement is high. In this situation, the structure of the YOLOv3 network is always redundant, and ABCP is proposed for the network pruning in simple tasks. Therefore, we evaluate ABCP on three detection datasets \cite{ABCP} including vehicle detection and mobile robot detection, whose classes are relatively simple, shown in Table \ref{tabd}.
%Therefore, we collect and label three relatively simple detection datasets \cite{ABCP} for ABCP evaluation, shown in Table \ref{tabd}.

\begin{table*}[t]
\renewcommand{\arraystretch}{1.3}
\caption{three datasets for experiments}\label{tabd}
\centering
\begin{tabular}{cc|c|c|c|c}
\hline
\multicolumn{2}{c|}{\multirow{2}{*}{Datasets}}                       & \multirow{2}{*}{resolution}                                               & \multicolumn{2}{c|}{frames} & \multirow{2}{*}{classes}                                                                            \\ \cline{4-5}
\multicolumn{2}{c|}{}                                                &                                                                           & training set  & testing set &                                                                                                     \\ \hline
\multicolumn{2}{c|}{UCSD dataset}                                    & 320$\times$240                                                            & 683           & 76          & truck, bus, car                                                                                     \\ \hline
\multicolumn{2}{c|}{mobile robot detection dataset}                  & \begin{tabular}[c]{@{}c@{}}1024$\times$512,\\ 640$\times$480\end{tabular} & 13,914        & 5,969       & \begin{tabular}[c]{@{}c@{}}red robot, red armor, blue robot, \\ blue armor, dead robot\end{tabular} \\ \hline
\multirow{2}{*}{sim2real detection dataset} & the simulation dataset & 640$\times$480                                                            & 5,760         & 1,440       & robot                                                                                               \\
                                            & the real-world dataset & 640$\times$480                                                            & ---             & 3,019       & robot                                                                                               \\ \hline
\end{tabular}
\end{table*}

\subsubsection{UCSD dataset}
\begin{figure}[t]
\centering
\subfloat[]{\includegraphics[width=1.1in]{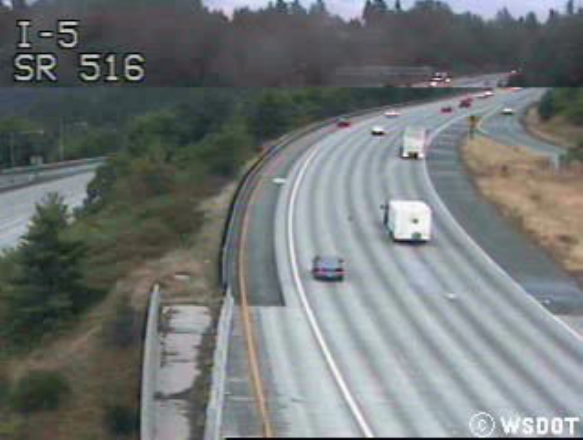}}
\hfil
\subfloat[]{\includegraphics[width=1.1in]{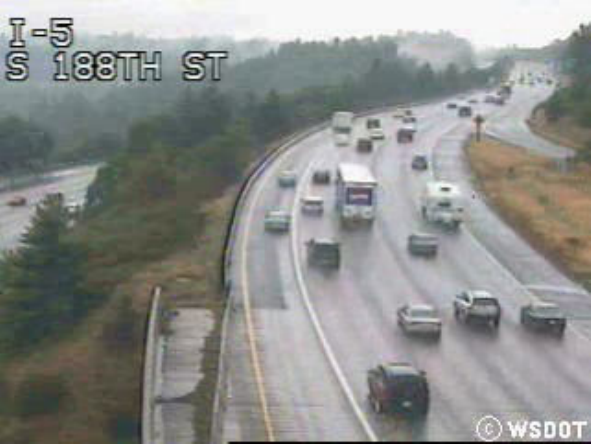}}
\hfil
\subfloat[]{\includegraphics[width=1.1in]{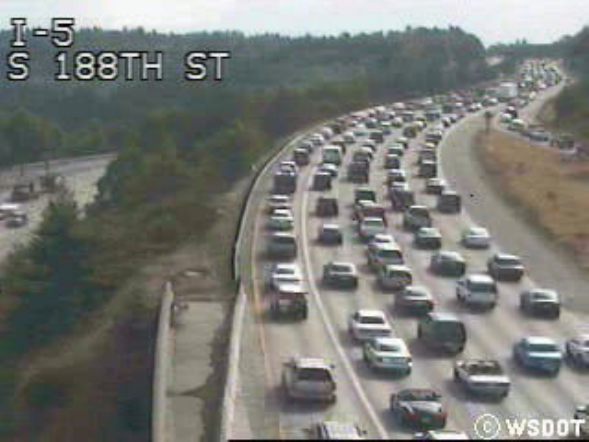}}
\caption{Examples of the UCSD dataset. (a) the sparse traffic; (b) the medium-density traffic; (c) the dense traffic.}
\label{fig3}
\end{figure}

The UCSD dataset is a small dataset captured from the freeway surveillance videos collected by UCSD \cite{ucsd}. As shown in Fig.\ref{fig3}, this dataset involves three different traffic densities each making up about one-third: the sparse traffic, the medium-density traffic, and the dense traffic. We define three classes in this dataset: truck, car, and bus. The resolutions of the images are all 320$\times$240. The training and testing sets contain 683 and 76 images respectively.

\subsubsection{Mobile robot detection dataset}
\begin{figure}[t]
\centering
\subfloat[]{\includegraphics[width=1.4in]{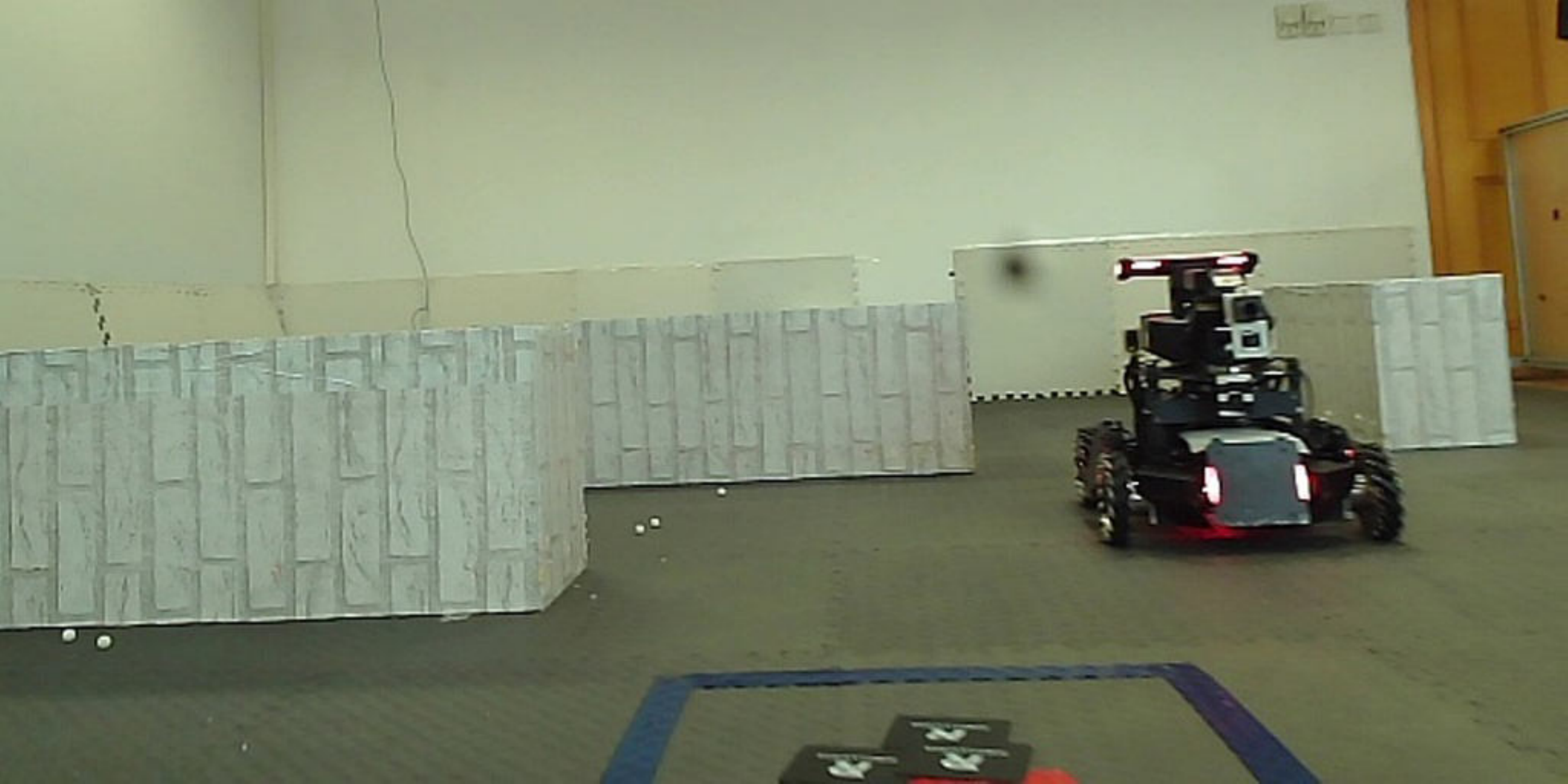}}
\hfil
\subfloat[]{\includegraphics[width=1.4in]{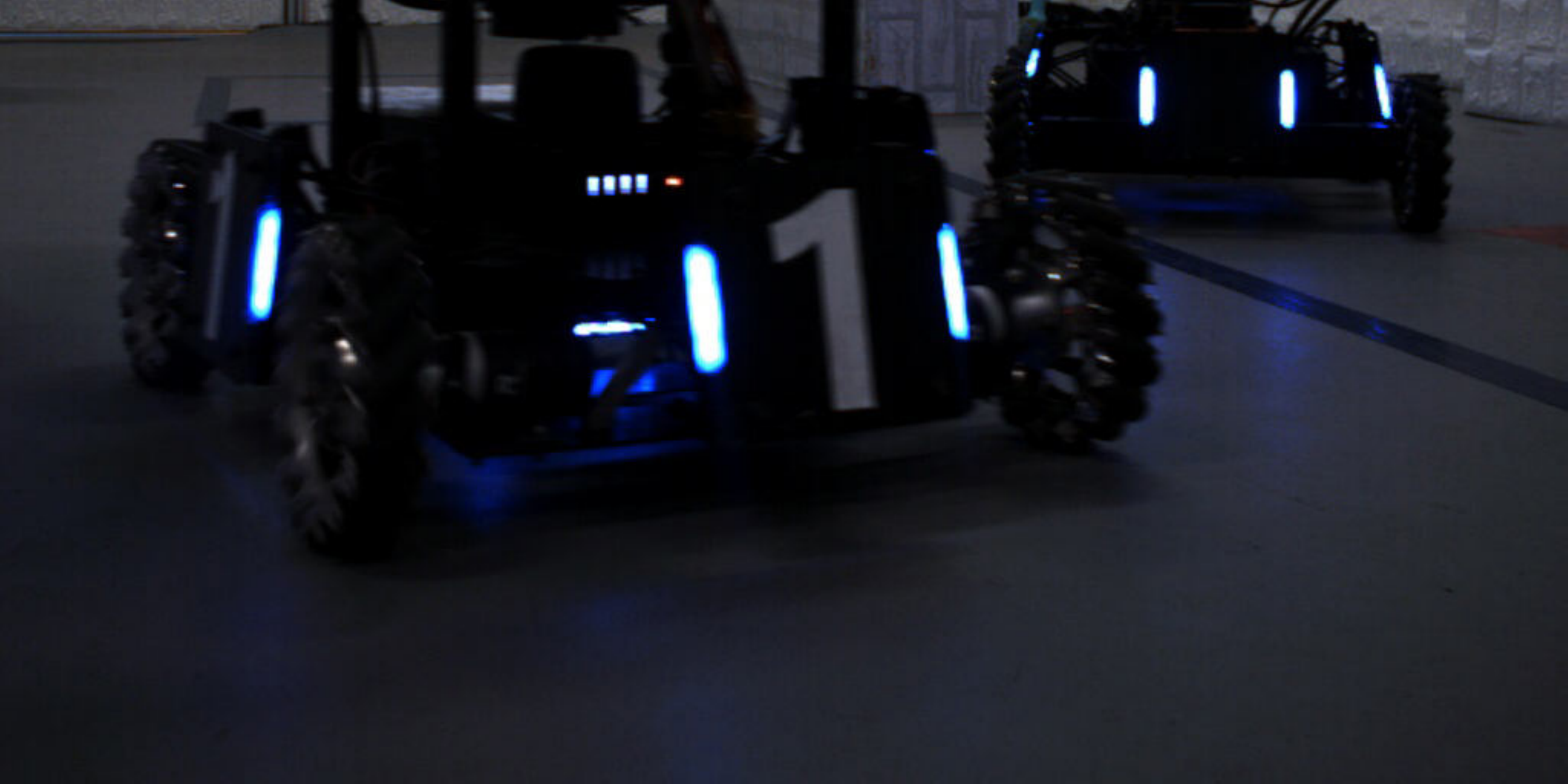}}
\hfil
\subfloat[]{\includegraphics[width=0.8in]{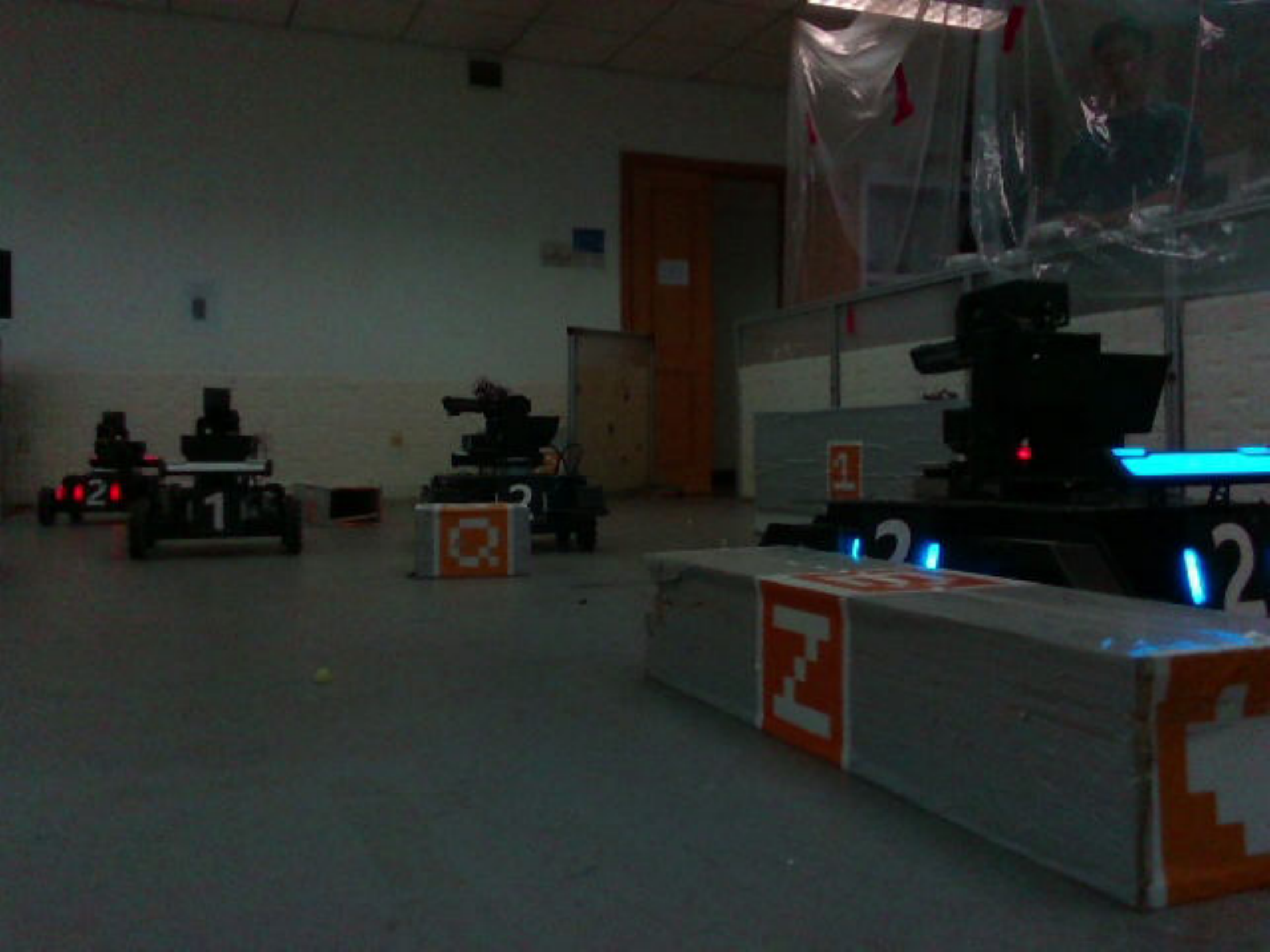}}
\hfil
\subfloat[]{\includegraphics[width=0.8in]{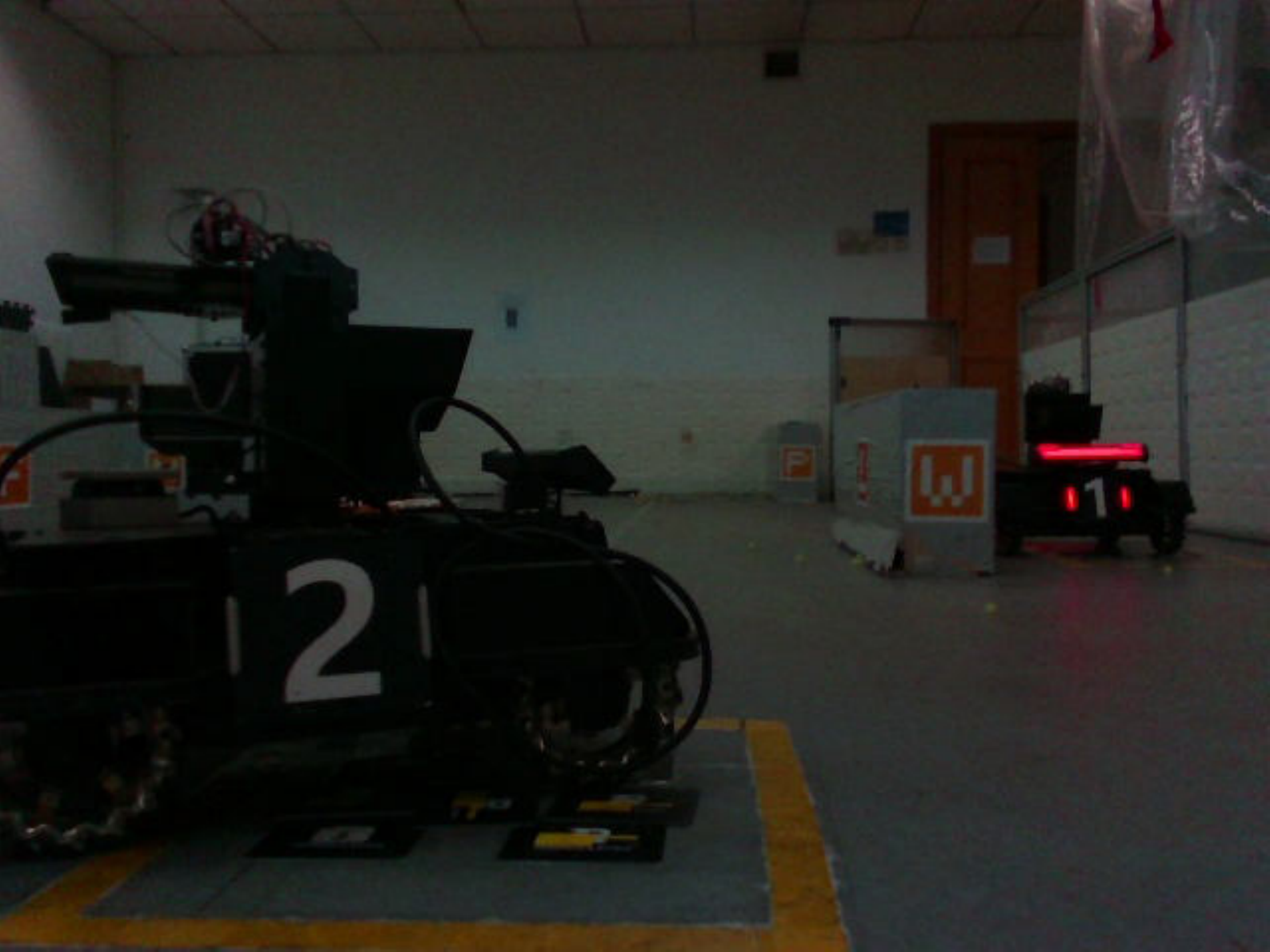}}
\hfil
\subfloat[]{\includegraphics[width=0.8in]{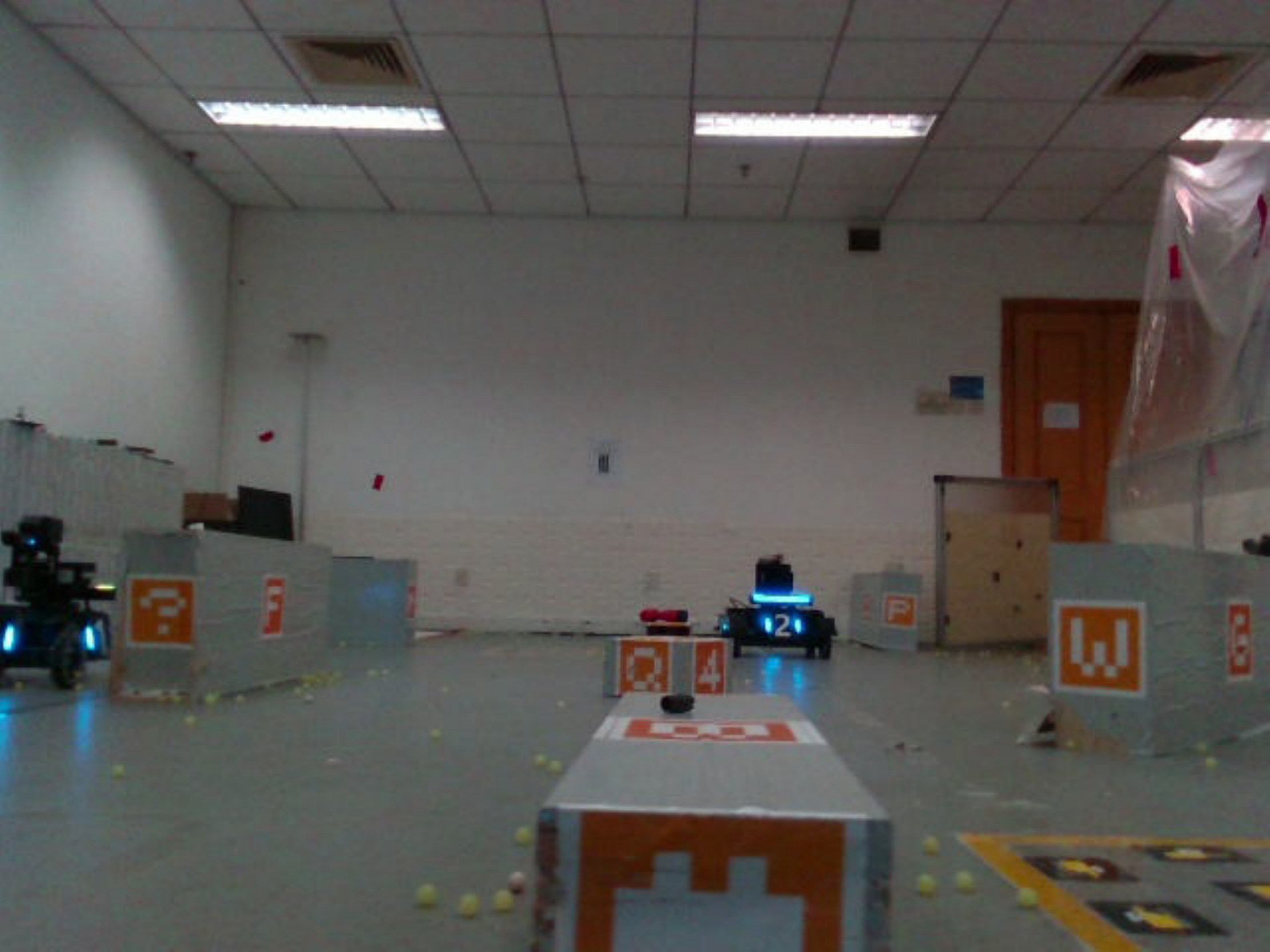}}
\hfil
\subfloat[]{\includegraphics[width=0.8in]{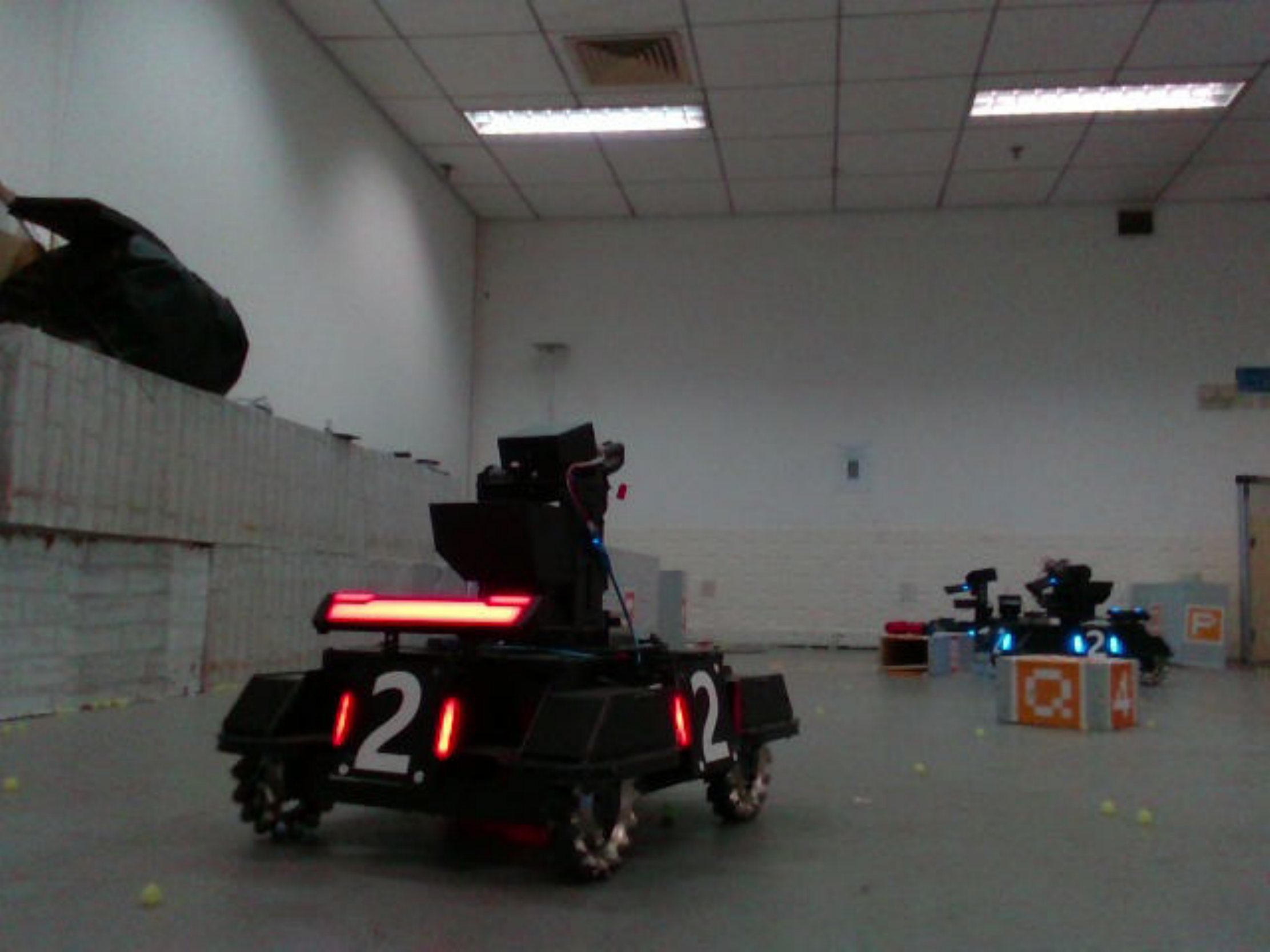}}
\caption{Examples of the mobile robot detection dataset. Different exposures as well as various distances and angles of the robots are performed. (a) and (b) are the 1024$\times$512 examples with different exposures, distances and angles; (c) - (f) are the 640$\times$480 examples with different exposures, distances and angles.}
\label{fig4}
\end{figure}

As shown in Fig.\ref{fig4}, the mobile robot detection dataset is collected by the robot-mounted cameras to meet the requirements of the fast and lightweight detection algorithms for the mobile robots. There are two kinds of ordinary color camera with different resolutions which are 1024$\times$512 and 640$\times$480 respectively. Five classes have been defined: red robot, red armor, blue robot, blue armor, dead robot. The training and testing sets contain $13,914$ and $5,969$ images respectively. During collecting, we change series of exposures as well as various distances and angles of the robots.

\subsubsection{Sim2real detection dataset}

\begin{figure}[t]
\centering
\subfloat[]{\includegraphics[width=1.3in]{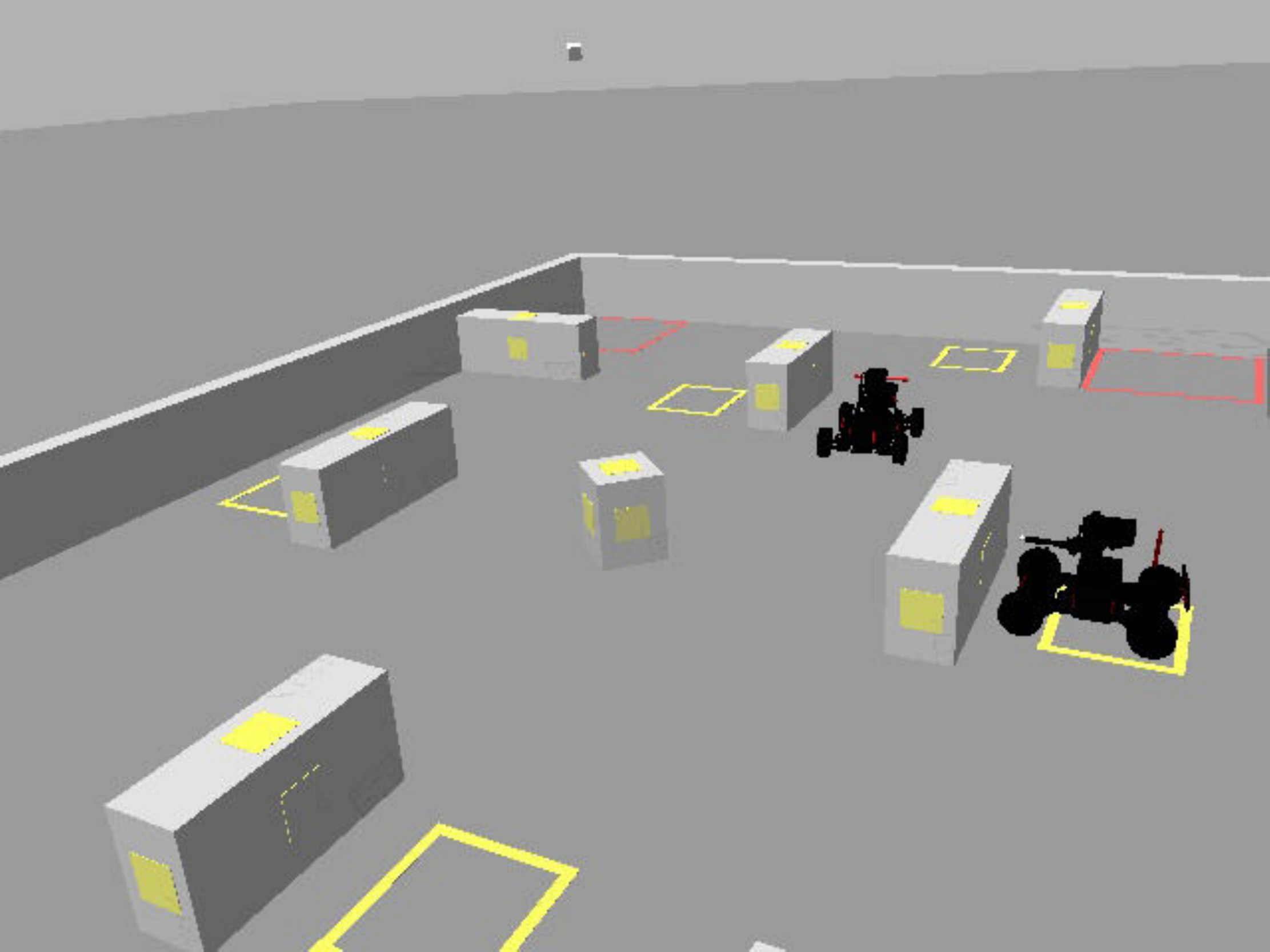}}
\hfil
\subfloat[]{\includegraphics[width=1.3in]{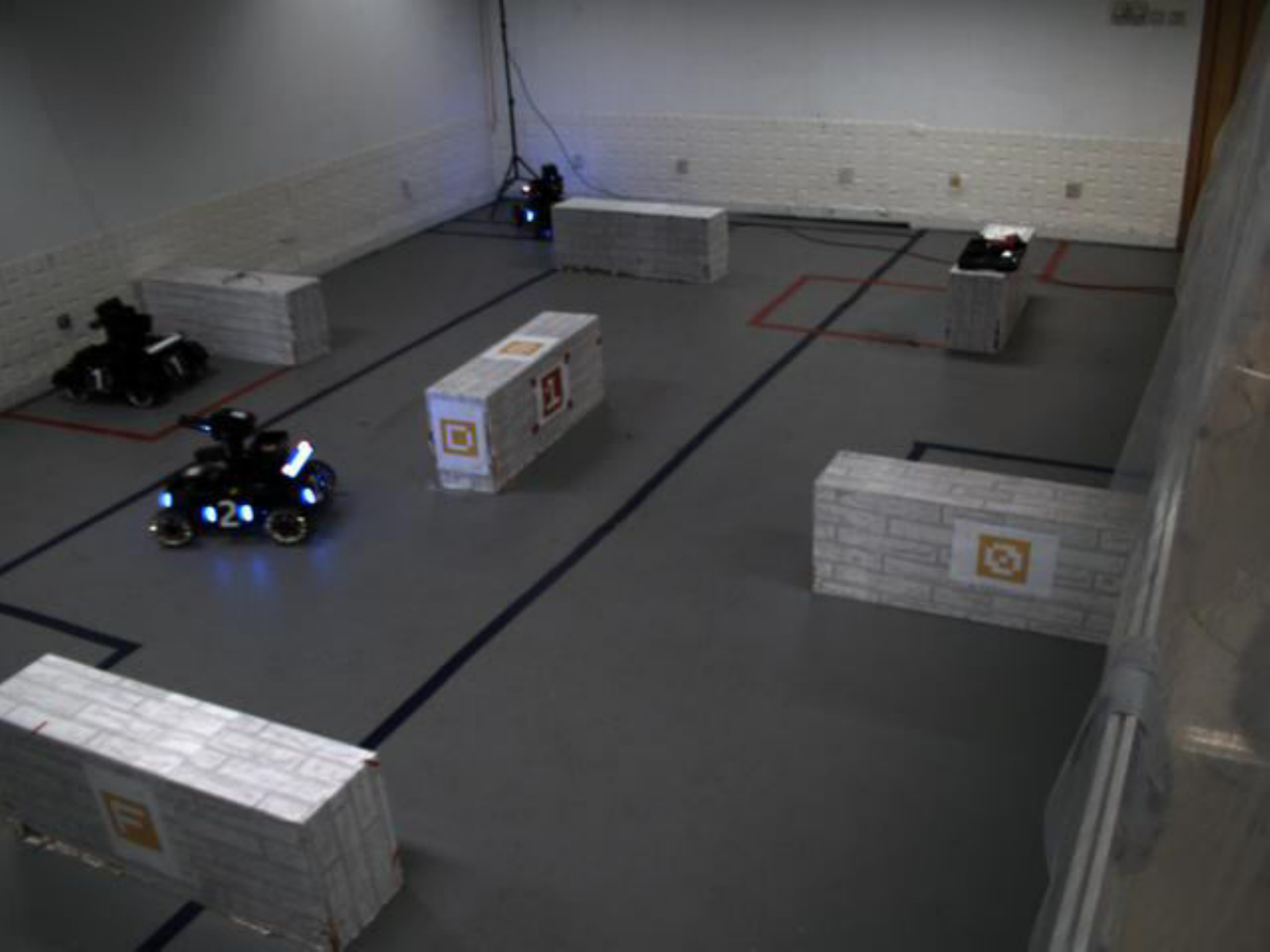}}
\caption{Examples of the sim2real detection dataset. (a) an example of the simulation dataset; (b) an example of the real-world dataset.}
\label{fig5}
\end{figure}

The sim2real detection dataset is divided into two sub-datasets: the real-world dataset and the simulation dataset. We search and train the model on the simulation dataset and test it on the real-world dataset. Firstly, we collect the real-world dataset by the surveillance-view ordinary color cameras in the field. The field and the mobile robots are the same as those in the mobile robot detection dataset. Secondly, we leverage Gazebo to simulate the robots and the field from the surveillance view. Then we capture the images of the simulation environment to collect the simulation dataset. The resolutions of images in the sim2real dataset are all 640$\times$480. There is only one object class in these two datasets: robot. The training and testing sets of the simulation dataset contain $5,760$ and $1,440$ images respectively, and the testing set of the real-world dataset contains $3,019$ images. Examples in the two datasets are demonstrated in Fig.\ref{fig5}.

\subsection{Implementation Details of ABCP}

We jointly search the block pruning choice with the discrete search space and the layer pruning ratio with the continuous search space in all of the experiments.

For pre-training the original YOLOv3 model, we firstly train the three prediction layers through the Adam optimizer for 20 epochs. The learning rate in the first stage is $10^{-3}$ for the UCSD dataset and is $10^{-4}$ for other datasets. Secondly, we train all of the weights in the model with Adam optimizer until the loss converges. The learning rate in the second stage is $10^{-4}$ for the UCSD dataset and is $10^{-6}$ for other datasets. The batch size is 64 and the weight decay is $5\times 10^{-4}$ in both two training stages. In addition, we perform the multi-scale image resizing and image augmenting to improve the accuracy.

For updating the parameters in the joint sample algorithm, we initialize the weights $\theta$ uniformly in $[-0.1, 0.1]$. For calculating the reward, $\lambda$ in \eqref{eq11} is set to $5 \times 10^5$ for the UCSD dataset and is set to $10^6$ for other datasets. The network of the joint sample algorithm is trained for 310 epochs with the Adam optimizer and the learning rate is $10^{-3}$. Thus the sample of the pruning action is also processed for 310 times. 
%Following \cite{enas}, the temperature is set to 5.0 and a tanh constant of 2.5 is applied to the outputs of the FC layers. In addition, the entropy of the joint sample algorithm is added to the reward, which is weighted by 0.1.

For fine-tuning the pruned model, we only train the three prediction layers in the pruned YOLOv3 for $1$ epoch with the Adam optimizer, with maintaining the values of the weights that have been set to zero after pruning. The learning rate in this stage is the same as that in the first stage of the pre-training, and the other sets are also the same as those in the pre-training.

For re-training, the final pruned model are re-trained by Darknet \cite{yolov3}. The optimizer is the Stochastic Gradient Descent (SGD) \cite{sgd}. The total batches are $30,000$ for the UCSD dataset and are $80,000$ for other datasets. The learning rate is set to $10^{-3}$ with no dropping. The batch size is set to $64$, the weight decay is $5\times 10^{-4}$ and the momentum factor is $0.9$.

\subsection{Compared Algorithms and Evaluation Metrics}

In the following experiments, we train the original YOLOv3 models \cite{yolov3} on the three datasets as the baseline and then prune the YOLOv3 models by ABCP. At present, YOLOv4 \cite{yolov4} develops a powerful method to improve YOLOv3 to get more efficient and more accurate. In addition, there is also a faster version of YOLOv3 named YOLO-tiny \cite{yolov3}. Hence, the YOLOv4 models and the YOLO-tiny models are also trained on the three datasets to compare with our pruned models. These models are all trained by Darknet with the SGD optimizer. During the training, following \cite{yolov3}, we set the total batches to $50,200$ for the YOLOv3 and YOLOv4 models, to $500,200$ for the YOLO-tiny models. The initial learning rate is $10^{-3}$, which drops to $10^{-4}$ at $80\%$ of the total batches and drops to $10^{-5}$ at $90\%$ of the total batches.

We also run a rule-based block-wise and channel-wise pruning algorithm (RBCP) proposed in \cite{Compression} to iteratively prune the YOLOv3 model. RBCP takes the YOLOv3 model trained by Darknet as the original model. During the iterative process, all the intermediate models and the finally pruned models are also trained by Darknet with the same hyper-parameters as those in the re-training of ABCP.

To evaluate the performance of the models, we use mean of average precision (mAP), FLOPs, the number
of the parameters (Params) and the average inference time to represent the accuracy, the complexity, and the inference speed of the models. During the evaluating, images are resized to 416$\times$416 before they are fed into the networks. The average inference time is tested on a NVIDIA MX250 GPU card, whose resource is very limited. In addition, there are three thresholds during the evaluation of the series of YOLO models \cite{yolov3}: the intersection over union (IOU) threshold is to calculate the IOUs between the predicted bounding boxes and the actual bounding boxes and to filter the predicted bounding boxes whose IOUs are smaller than the IOU threshold, which is set to $0.5$; the confidence threshold is to filter the predicted bounding boxes whose confidences are smaller than the confidence threshold, which is set to $0.8$; and the non-maximum suppression threshold \cite{nms} is set to $0.5$.

\subsection{Ablation Study}

We ascribe the excellent performance of ABCP to two points: 1) ABCP prunes both residual blocks and channels via the joint search of the block-wise and channel-wise pruning action; 2) the search space for channel pruning is continuous. In the following experiments shown in Table \ref{tab4}, we prove the contributions of these two points on the UCSD dataset. 

\subsubsection{Effects of the joint search}
To explore the effectiveness of the block-wise and channel-wise joint search, as shown in Table \ref{tab4}, the model pruned by ABCP is compared with two models pruned through the single-wise search: ABCP-w/o-C is the model pruned with the action generated by the block-wise pruning action search; ABCP-w/o-B is the model pruned with the action generated by the channel-wise pruning action search. The implementation details are the same as those of ABCP. 

The results show that compared with the FLOPs of ABCP, ABCP-w/o-B can achieve a comparable compression ratio while ABCP-w/o-C is still much more resource-consuming. It is because that the block-wise pruning is coarse-grained while the channel-wise pruning can prune more fine-grained structures. Therefore, ABCP combines the coarse-grained and fine-grained pruning and can obtain an ultra-slim pruned model. As for the comparison of accuracy, ABCP reaches the highest mAP among these models. It is validated that the joint search can sample a better pruning action to accomplish a larger compression ratio with low accuracy loss.

\subsubsection{Effects of the continuous search space}
To check the effects of the continuous search space for the channel-wise pruning, as shown in Table \ref{tab4} (ABCP-D), we prune the model with the action generated by the joint search, while the action space is discrete for the search of the channel-wise pruning, which is set to ${\{0, 0.225, 0.45, 0.675, 0.9\}}$. The implementation details are the same as those of ABCP. 

The results show that ABCP achieves better accuracy with fewer FLOPs as well as the comparable Params and inference speed. It is verified by experiments that the continuous search space is more suitable for the pruning task since the models are sensitive for the layers to be pruned by different pruning ratios. In addition, the continuous search space contributes to getting a higher compression ratio.

\begin{table}[t]
\renewcommand{\arraystretch}{1.3}
\caption{results of the ablation study on the UCSD dataset}\label{tab4}
\centering
\begin{tabular}{c|c|l|c|c|c}
\hline
\multirow{2}{*}{Models} & \multicolumn{2}{c|}{\multirow{2}{*}{\begin{tabular}[c]{@{}c@{}}mAP\\ (\%)\end{tabular}}} & \multirow{2}{*}{\begin{tabular}[c]{@{}c@{}}FLOPs\\ (G)\end{tabular}} & \multirow{2}{*}{\begin{tabular}[c]{@{}c@{}}Params\\ (M)\end{tabular}} & \multirow{2}{*}{\begin{tabular}[c]{@{}c@{}}Inference\\ Time (s)\end{tabular}} \\
                        & \multicolumn{2}{c|}{}                                                                    &                                                                      &                                                                       &                                                                               \\ \hline
ABCP                    & \multicolumn{2}{c|}{\textbf{69.6}}                                                                & \textbf{4.485}                                                                & 4.685                                                                 & 0.016                                                                        \\
ABCP-w/o-C                  & \multicolumn{2}{c|}{64.3}                                                                & 35.283                                                               & 36.096                                                                & 0.061                                                                         \\
ABCP-w/o-B                  & \multicolumn{2}{c|}{58.1}                                                                & 4.943                                                                & 9.249                                                                 & 0.020                                                                         \\
ABCP-D                  & \multicolumn{2}{c|}{63.5}                                                                & 4.924                                                                & \textbf{4.340}                                                                 & \textbf{0.015}                                                                         \\ \hline
\end{tabular}
\end{table}

\subsection{Results and Analysis}

According to the conclusion of the ablation study, we train and prune YOLOv3 models on three datasets by ABCP. Then the pruned models are compared with YOLOv4, YOLO-tiny, as well as the models pruned by RBCP.

\subsubsection{Results on the UCSD dataset}

\begin{figure}[t]
    \centering
    \includegraphics[width=2.2in]{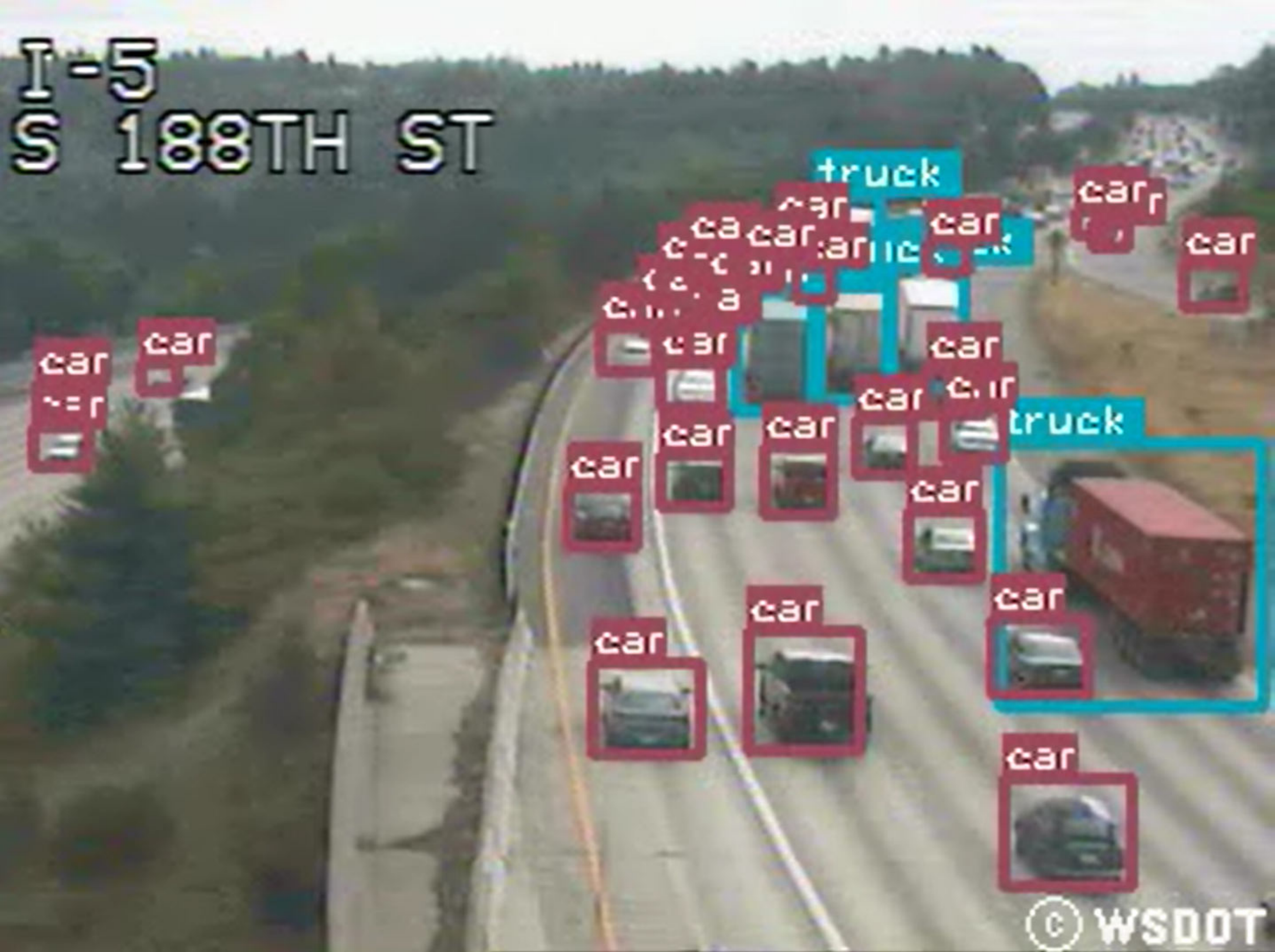}
    \caption{The detection results of the model pruned by ABCP on the UCSD dataset.}
    \label{fig6}
    \end{figure}

The performances of the models on the UCSD dataset are demonstrated in Table \ref{tab1}. The results show that the mAP of our pruned model surpasses the baseline YOLOv3 model by $\bm{8.2\%}$ with $\bm{93.2\%}$ FLOPs reduction, $\bm{92.4\%}$ Params reduction, and $\bm{6.87\times}$ speed up. Fig. \ref{fig6} shows the great detection results of our pruned model, and the video of the detection results is demonstrated on the github. The pruned model also achieves 6.5\% higher mAP than the YOLOv4 model with much fewer FLOPs and Params as well as much faster speed. The highest mAP of ABCP reflects the redundancy of the structures of YOLOv3 and YOLOv4, which are not suitable for the relatively simple detection tasks. Compared with the YOLO-tiny model, ABCP outperforms it by a large margin. It is due to in YOLO-tiny, the parameters are reduced by replacing the residual blocks with the ordinary convolutional layers and cutting a level of the feature pyramid structure, leading to irrecoverable accuracy loss. 

As for the rule-based block-wise and channel-wise joint pruning method RBCP, since the pruning process of RBCP is iterative, we perform the pruning pipeline iteratively until the pruning causes a dramatic accuracy loss or the results of the sparse training indicates that there are almost no redundant structures to be pruned. %Fig. \ref{fig7} illustrates the pruning iteration of RBCP.The accuracy of the pruned model first decreases and then increases, and eventually surpasses the accuracy of the baseline model, which indicates that the structure of the YOLOv3 model is redundant for this task and the pruning process of RBCP is to search for the best structure step by step. As shown in Fig. \ref{fig7}, 
During the pruning iteration of RBCP, the mAP of the 7th pruned model reaches 66.5\%, but the FLOPs and Params are both much larger than ours. However, in the next iteration, the mAP of the 8th pruned model drops significantly, while the FLOPs reduction is little. In addition, after the 8th iteration, almost no parameter is close to zero after sparse learning, the iteration process is forced to terminate. Hence, we terminate the pruning iteration here and take the performance of the 7th pruned model as the result of RBCP in Table. \ref{tab1}. Compared with the performance of the models pruned by ABCP shown in Table. \ref{tab1}, it is verified by experiments that RBCP can not achieve the sufficient compression ratio.
%To add to this, Fig. \ref{fig7} also shows that ABCP has the best performance among the three compression methods.

%\begin{figure}[t]
%    \centering
%    \includegraphics[width=2.4in]{fig/star.pdf}
%    \caption{The performances of the models compressed by three different methods.}
%    \label{fig7}
%\end{figure}

\begin{table}[t]
\renewcommand{\arraystretch}{1.3}
\caption{results of the models on the UCSD dataset}\label{tab1}
\centering
\begin{tabular}{c|c|c|c|c}
\hline
Models               & \begin{tabular}[c]{@{}c@{}}mAP\\ (\%)\end{tabular} & \begin{tabular}[c]{@{}c@{}}FLOPs\\ (G)\end{tabular} & \begin{tabular}[c]{@{}c@{}}Params\\ (M)\end{tabular} & \begin{tabular}[c]{@{}c@{}}Inference\\ Time (s)\end{tabular} \\ \hline
YOLOv3 \cite{yolov3}               & 61.4                                               & 65.496                                              & 61.535                                               & 0.110                                                        \\
YOLOv4 \cite{yolov4}               & 63.1                                               & 59.659                                              & 63.948                                               & 0.132                                                        \\
YOLO-tiny \cite{yolov3}           & 57.4                                               & 5.475                                               & 8.674                                                & \textbf{0.014}                                                        \\
RBCP \cite{Compression} & 66.5                                              & 17.973                                              & 4.844                                                & 0.042                                                        \\
ABCP (Ours)                 & \textbf{69.6}                                               & \textbf{4.485}                                              & \textbf{4.685}                                                & 0.016                                                        \\ \hline
\end{tabular}
\end{table}

\begin{figure*}[t]
\centering
\includegraphics[width=7in]{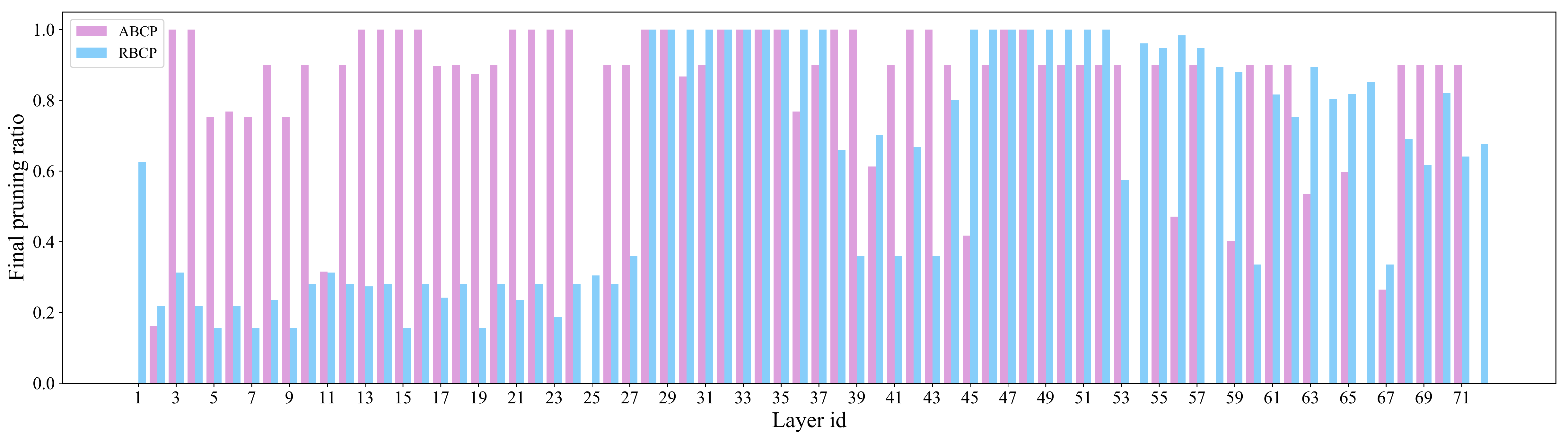}
\caption{The pruning ratios of each layer of the models with the best performance pruned by ABCP and RBCP. The vacancy of the column represents the pruning ratio is 0\%.}
\label{fig8}
\end{figure*}

Additionally, Fig. \ref{fig8} demonstrates the pruning ratios of each layer of the models with the best performance pruned by RBCP and ABCP. The biggest difference between the two policies is the pruning ratios of the first 24 layers. For RBCP, the pruning ratio of each layer is the sum of the pruning ratios generated in all previous iterations. Hence, during the pruning iteration, the pruning ratios of the first 24 layers at each time are always confined in a small range, which results in the limited FLOPs reduction of RBCP. We suppose that it is another manifestation of the limited compression of RBCP. Therefore, it is easier for ABCP to find the best pruned network.

Furthermore, we attempt to use RBCP to prune the model pruned by ABCP. The block-wise pruning and the channel-wise pruning are processed iteratively in RBCP. After the block-wise pruning, three residual blocks are pruned and the FLOPs is reduced by 0.07G. Nevertheless, the inference speed is the same (0.016s) and the mAP is dropped by 0.04\%. Sequentially, the channel-wise pruning is performed, but the results of the sparse training demonstrate that there are almost no redundant channels in the model. It is verified by experiments that the model pruned by ABCP can no longer be optimized by RBCP.

\subsubsection{Results on the mobile robot detection dataset}

\begin{figure}[t]
\centering
\includegraphics[width=2.4in]{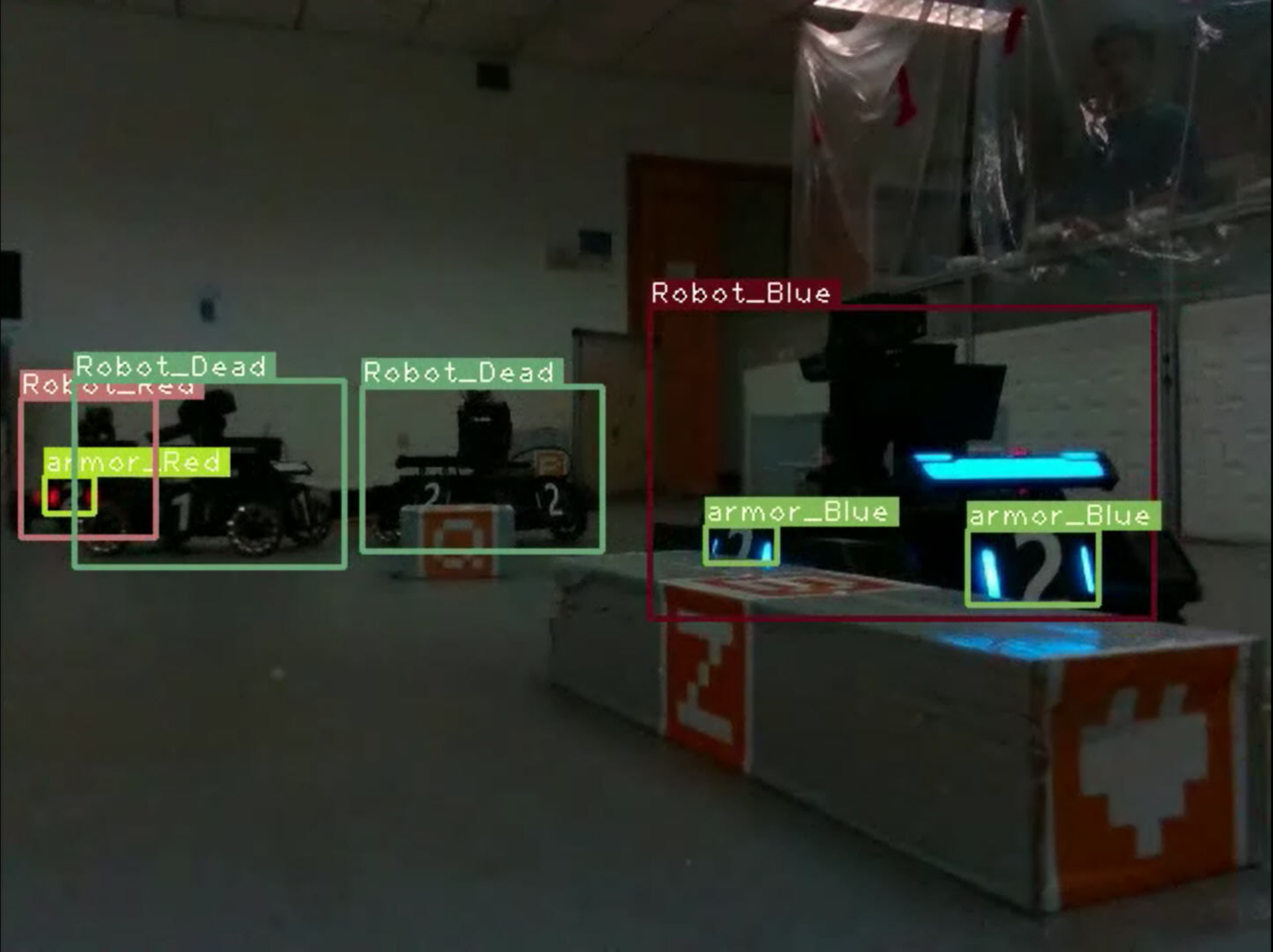}
\caption{The detection results of the model pruned by ABCP on the mobile robot detection dataset.}
\label{fig9}
\end{figure}

The performances of the models on the mobile robot detection dataset are demonstrated in Table \ref{tab2}. The results show that compared with the baseline YOLOv3 model, our pruned model saves $\bm{99.5\%}$ FLOPs, reduces $\bm{99.5\%}$ Params, and achieves $\bm{37.3\times}$ speed up with only 2.8\% accuracy loss. The accuracy of our pruned model is the same as that of YOLOv4. It is also demonstrated that reaching excellent detection accuracy on the simple task does not require such complex structures as YOLOv3 and YOLOv4. Fig. \ref{fig9} shows the great detection results of our pruned model, and the video of the detection results is demonstrated on the github. Compared with YOLO-tiny and RBCP, ABCP outperforms them by a large margin with much lower accuracy loss and much higher compression ratio. 

Compared with the UCSD dataset, the objects in the mobile robot detection dataset are sparser and more obvious, so that the detection task on this dataset is simpler. Hence, the results also illustrate that ABCP can work much better than RBCP on a simple dataset. 

Moreover, we deploy the model pruned by ABCP on the NVIDIA$^{\circledR}$ Jetson AGX Xavier$^\text{TM}$ platform mounted on the robot. After speeded up by NVIDIA$^{\circledR}$ TensorRT$^\text{TM}$, the inference speed can reach approximate 300 frames per second, which manifests that the pruned model also has remarkable property on the embedded devices. 

\begin{table}[t]
\renewcommand{\arraystretch}{1.3}
\caption{results of the models on the mobile robot detection dataset}\label{tab2}
\centering
\begin{tabular}{c|c|c|c|c}
\hline
Models             & \begin{tabular}[c]{@{}c@{}}mAP\\ (\%)\end{tabular} & \begin{tabular}[c]{@{}c@{}}FLOPs\\ (G)\end{tabular} & \begin{tabular}[c]{@{}c@{}}Params\\ (M)\end{tabular} & \begin{tabular}[c]{@{}c@{}}Inference\\ Time (s)\end{tabular} \\ \hline
YOLOv3 \cite{yolov3}             & \textbf{94.9}                                               & 65.510                                              & 61.545                                               & 0.112                                                        \\
YOLOv4 \cite{yolov4}           & 92.1                                               & 59.673                                              & 63.959                                               & 0.141                                                        \\
YOLO-tiny \cite{yolov3}         & 85.3                                               & 5.478                                               & 8.679                                                & 0.014                                                        \\
RBCP \cite{Compression} & 89.9                                               & 2.842                                               & 1.879                                                & 0.012                                                        \\
ABCP (Ours)               & 92.1                                               & \textbf{0.327}                                               & \textbf{0.299}                                                & \textbf{0.003}                                                        \\ \hline
\end{tabular}
\end{table}

\subsubsection{Results of the transfer task on the sim2real detection dataset}

\begin{figure*}[t]
    \begin{minipage}{0.1\textwidth}
    \centering{\bf YOLOv3 \cite{yolov3}}
    \end{minipage}
    \begin{minipage}{0.9\textwidth}
    \includegraphics[width=\textwidth]{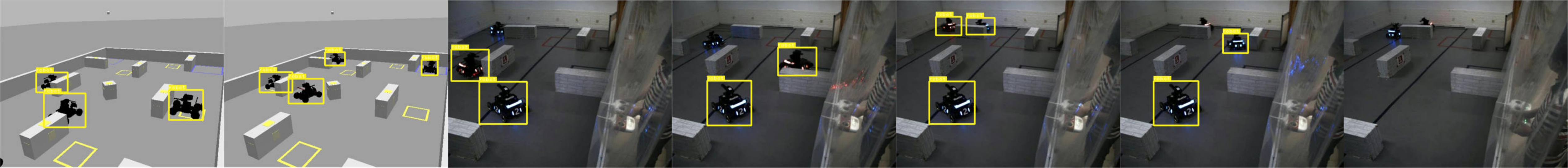}
    \end{minipage}

    \begin{minipage}{0.1\textwidth}
    \centering{\bf YOLOv4 \cite{yolov4}}
    \end{minipage}
    \begin{minipage}{0.9\textwidth}
    \includegraphics[width=\textwidth]{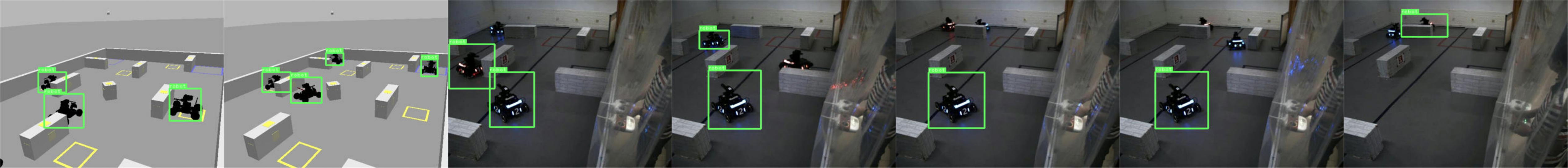}
    \end{minipage}

    \begin{minipage}{0.1\textwidth}
    \centering{\bf YOLO-tiny \cite{yolov3}}
    \end{minipage}
    \begin{minipage}{0.9\textwidth}
    \includegraphics[width=\textwidth]{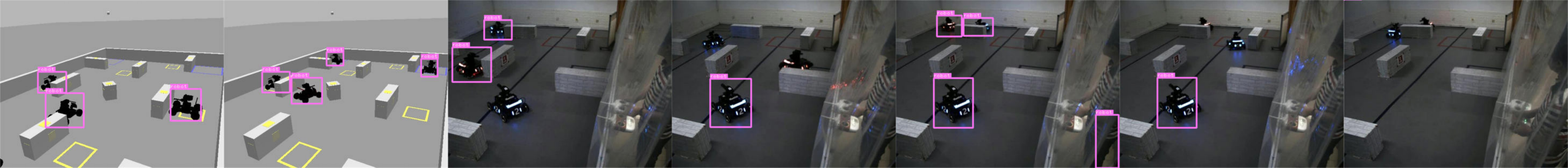}
    \end{minipage}

    \begin{minipage}{0.1\textwidth}
    \centering{\bf RBCP \cite{Compression}}
    \end{minipage}
    \begin{minipage}{0.9\textwidth}
    \includegraphics[width=\textwidth]{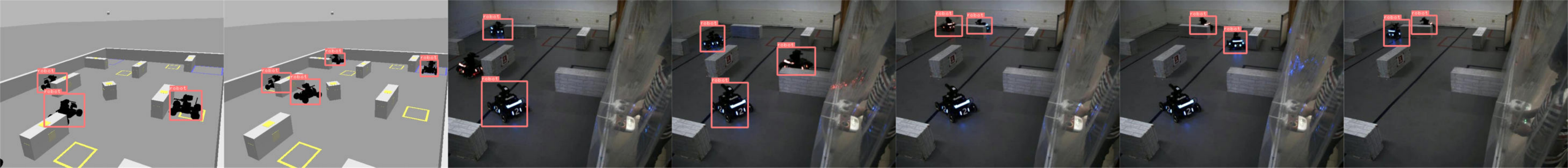}
    \end{minipage}

    \begin{minipage}{0.1\textwidth}
    \centering{\bf ABCP (Ours)}
    \end{minipage}
    \begin{minipage}{0.9\textwidth}
    \includegraphics[width=\textwidth]{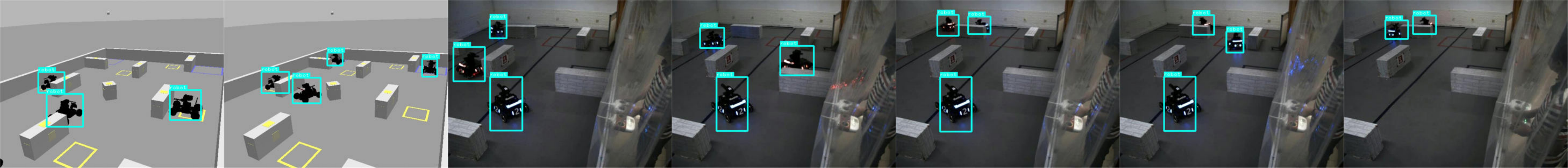}
    \end{minipage}
\caption{The visualization of the detection results comparison on the simulation dataset and the real-world dataset. The first two columns are the detection results on the simulation dataset, and other columns are the detection results on the real-world dataset.}
\label{fig10}
\end{figure*}

The transfer task on the sim2real dataset is to train the model on the simulation dataset and then transfer the model on the real-world dataset. In the following experiments, we search and train the model on the simulation dataset, and directly transfer the weights to the real-world dataset with no fine-tuning to evaluate the performance on the real-world dataset. Table \ref{tab3} illustrates the results of the models on this task. Compared with the baseline YOLOv3 model, under $\bm{97.6\%}$ FLOPs reduction, $\bm{95.8\%}$ Params reduction, and $\bm{14.6\times}$ speed up, our pruned model achieves $\bm{2.4\%}$ better accuracy on the simulation dataset and $\bm{9.6\%}$ better accuracy on the real-world dataset. Compared with other models, the performances tested on the simulation dataset are similar, while the model pruned by ABCP achieves the best accuracy on the real-world dataset with the fewest FLOPs. Fig. \ref{fig10} shows the visualization of the detection results comparison on the simulation dataset and the real-world dataset, and the video of the detection results of the model pruned by ABCP is demonstrated on the github. It can be seen that the model pruned by ABCP has better accuracy on the real-world dataset. It is verified by experiments that the model pruned by ABCP has better robustness performance. 

Furthermore, it has been shown that YOLOv4 does not perform well on this transfer task, which may be caused by the overfitting problems as the YOLOv4 model has more redundant parameters. At the same time, YOLOv3 also has the overfitting problems. In addition, the accuracy of YOLO-tiny on the real-world dataset loses much more than ABCP, probably reflecting that the slim model generated by this method does not capture all of the available features of the objects.

\begin{table}[t]
\renewcommand{\arraystretch}{1.3}
\caption{Results of the transfer task on the sim2real detection dataset}\label{tab3}
\centering
\begin{tabular}{c|c|c|c|c|c}
\hline
\multirow{2}{*}{Models} & \multicolumn{2}{c|}{mAP (\%)}                                                                                            & \multirow{2}{*}{\begin{tabular}[c]{@{}c@{}}FLOPs\\ (G)\end{tabular}} & \multirow{2}{*}{\begin{tabular}[c]{@{}c@{}}Params\\ (M)\end{tabular}} & \multirow{2}{*}{\begin{tabular}[c]{@{}c@{}}Inference\\ Time (s)\end{tabular}} \\ \cline{2-3}
                        & \begin{tabular}[c]{@{}c@{}}sim\\dataset\end{tabular} & \begin{tabular}[c]{@{}c@{}}real\\dataset\end{tabular} &                                                                      &                                                                       &                                                                               \\ \hline
YOLOv3 \cite{yolov3}                 & 95.6                                                         & 66.5                                                     & 65.481                                                               & 61.524                                                                & 0.117                                                                         \\
YOLOv4 \cite{yolov4}                 & \textbf{98.3}                                                         & 28.8                                                     & 59.644                                                               & 63.938                                                                & 0.141                                                                         \\
YOLO-tiny \cite{yolov3}              & \textbf{98.3}                                                         & 42.3                                                     & 5.472                                                                & 8.670                                                                 & 0.014                                                                         \\
RBCP \cite{Compression}                   & 97.9                                                         & 71.2                                                     & 2.321                                                                & \textbf{1.237}                                                                 & 0.009                                                                         \\
ABCP (Ours)                   & 98.0                                                         & \textbf{76.1}                                                     & \textbf{1.581}                                                                & 2.545                                                                 & \textbf{0.008}                                                                         \\ \hline
\end{tabular}
\end{table}

\section{Conclusion}

In this paper, we propose ABCP, which jointly search the block-wise and channel-wise pruning action through DRL, pruning both residual blocks and channels automatically. A joint sample algorithm is proposed to generate the pruning choice of each residual block and the channel pruning ratio of each convolutional layer in the models. Evaluated on YOLOv3 with three datasets, the results indicate that our method outperforms the traditional rule-based pruning methods with better accuracy and higher compression ratio. Furthermore, since ABCP is only applied to the detection models in the experiments, we would like to apply the proposed method on more deep learning tasks, such as image classification and 3D object detection.

%\section*{Acknowledgment}
%The authors would like to thank the research partners Zixiang Ding and Nannan Li of the State Key Laboratory of Management  and Control for Complex Systems, Institute of Automation, Chinese Academy of Sciences, for their help through each stage of the research process.

{
\bibliographystyle{IEEEtran}
\bibliography{IEEEroot}
}

\end{document}